\documentclass[letterpaper, 10 pt, conference]{ieeeconf}  % Comment this line out if you need a4paper
 \usepackage{amsmath} 
 \usepackage{graphicx} 
 \usepackage{graphicx}
\usepackage{hyperref}
\usepackage{amssymb}
\usepackage{algorithm}
\usepackage{algpseudocode}
\usepackage{amsmath}
\usepackage{amsfonts}
\usepackage{url}
\usepackage{textcomp}
\usepackage{stfloats}
\usepackage{multirow}
\usepackage{wrapfig}
\usepackage{caption}
\usepackage{stackengine}
\usepackage{subcaption}
\usepackage{xcolor,cite}
\usepackage{tabu}                     % 表格插入
\usepackage{multirow}                 % 一般用以设计表格，将所行合并
\usepackage{multicol}                 % 合并多列
\usepackage{multirow}                % 合并多行
\usepackage{float}                    % 图片浮动
\usepackage{makecell}                 % 三线表-竖线
\usepackage{booktabs}   
\usepackage[table]{xcolor}
\usepackage{nicematrix}
\usepackage{nccmath} % 引入 nccmath 包

\IEEEoverridecommandlockouts                              % This command is only needed if 
\title{\LARGE \bf
AugVLA-3D: Depth-Driven Feature Augmentation for Vision-Language-Action Models
}

\author{
Zhifeng Rao$^{1,2*}$,
Wenlong Chen$^{1*}$,
Lei Xie$^{1}$,
Xia Hua$^{3}$,
Dongfu Yin$^{1}$,
Zhen Tian$^{1\dagger}$,
F. Richard Yu$^{4}$%
\thanks{*These authors contributed equally to this work.}%
\thanks{$\dagger$ Corresponding author: Zhen Tian, {\tt\small tianzhen@gml.ac.cn}.}%
\thanks{This work was supported by the Guangdong Natural Science Foundation (2025A1515012083) and the National Natural Science Foundation of China (62271324, 62231020).}%
\thanks{$^{1}$Guangdong Laboratory of Artificial Intelligence and Digital Economy (SZ), Shenzhen, China.}%
\thanks{$^{2}$Department of Computer Science and Engineering, Southern University of Science and Technology, Shenzhen, China.}%
\thanks{$^{3}$School of Communication and Information Engineering, Shanghai University, Shanghai, China.}%
\thanks{$^{4}$School of Information Technology, Carleton University, Ottawa, Canada.}%
}

\begin{document}

\maketitle
\thispagestyle{empty}
\pagestyle{empty}

%%%%%%%%%%%%%%%%%%%%%%%%%%%%%%%%%%%%%%%%%%%%%%%%%%%%%%%%%%%%%%%%%%%%%%%%%%%%%%%%
\begin{abstract}
Vision-Language-Action (VLA) models have recently achieved remarkable progress in robotic perception and control, yet most existing approaches primarily rely on VLM trained using 2D images, which limits their spatial understanding and action grounding in complex 3D environments. To address this limitation, we propose a novel framework that integrates depth estimation into VLA models to enrich 3D feature representations. Specifically, we employ a depth estimation baseline called VGGT to extract geometry-aware 3D cues from standard RGB inputs, enabling efficient utilization of existing large-scale 2D datasets while implicitly recovering 3D structural information. To further enhance the reliability of these depth-derived features, we introduce a new module called action assistant, which constrains the learned 3D representations with action priors and ensures their consistency with downstream control tasks. By fusing the enhanced 3D features with conventional 2D visual tokens, our approach significantly improves the generalization ability and robustness of VLA models. Experimental results demonstrate that the proposed method not only strengthens perception in geometrically ambiguous scenarios but also leads to superior action prediction accuracy. This work highlights the potential of depth-driven data augmentation and auxiliary expert supervision for bridging the gap between 2D observations and 3D-aware decision-making in robotic systems.

\end{abstract}

%%%%%%%%%%%%%%%%%%%%%%%%%%%%%%%%%%%%%%%%%%%%%%%%%%%%%%%%%%%%%%%%%%%%%%%%%%%%%%%%
\section{Introduction}
A central ambition in robotics is to develop agents with broad versatility: the capability to execute diverse tasks across varying environments while following natural language instructions, adapting to situational constraints, and remaining resilient to unexpected disturbances. Yet, attaining such generality is still highly challenging due to the inherent complexity of real-world perception–action loops and the wide variability of manipulation scenarios. Recent advances in imitation learning \cite{Linetal2024,lin2024learning,jiang2024dexmimicgen,ren2025motion,liu2025forcemimic} together with the rise of Vision-Language-Action (VLA) models \cite{zhen20243d,zhao2025cot,chiang2024mobility,yue2024deer,ding2024quar} provide a promising route toward building robots with adaptable and transferable skills. By combining large-scale vision-language models (VLMs) with action-generation modules, VLAs have created new opportunities for scalable and generalizable policy learning.
\begin{figure*}[t]
\begin{center}
\hypertarget{figure:figure1}{}
\label{figure:figure1}
\centering
 \includegraphics[width=14cm,height=9.5cm]{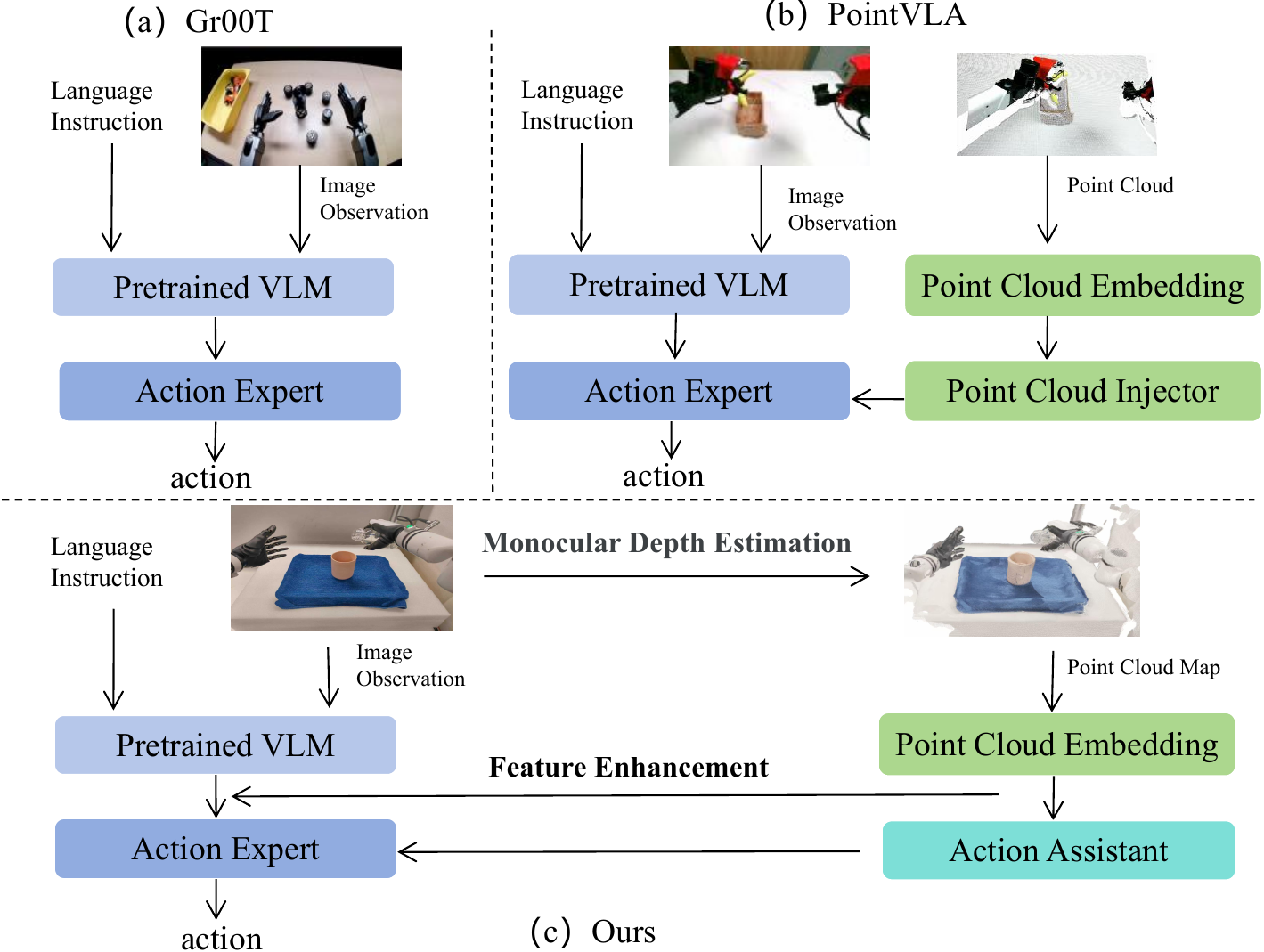}
\end{center}
\caption{The architecture comparison with different methods. (a) Gr00t \cite{bjorck2025gr00t}: Only 2D visual features are used without explicit 3D reasoning. 
(b) PointVLA \cite{li2025pointvla}: LiDAR-based point clouds are introduced but rely on specialized 3D sensors. (c) AugVLA-3D: Our AugVLA-3D leverages depth estimation to inject 3D structural features in a sensor-free manner, enabling scalable training and stronger 3D generalization.}
\end{figure*}

Recent years have witnessed remarkable progress in this line of research. OpenVLA \cite{kim2024openvla} demonstrates that scaling vision-language pretraining to billions of parameters enables robots to generalize across a wide range of manipulation tasks. Building on this foundation, $\pi$0 \cite{black2410pi0} leverages pretrained VLMs to align visual observations with textual task descriptions, thereby enabling flexible policy learning from human demonstrations. Its successor, $\pi$0.5 \cite{black2025pi0}, further refines multimodal alignment strategies and incorporates broader datasets, leading to improved task generalization and grounding. Despite these advances, a key limitation remains: these models fundamentally rely on 2D representations inherited from VLMs. While 2D features capture rich semantic context, they lack explicit 3D structural awareness that is indispensable for reasoning about geometry, depth, and spatial interactions in the physical world. Consequently, such models often falter in scenarios requiring fine-grained spatial reasoning, such as collision avoidance, object stacking, or reachability analysis.

To overcome the shortcomings of purely 2D approaches, recent works have attempted to introduce explicit 3D information. 3D-VLA \cite{zhen20243d} proposes a generative world model that integrates 3D perception, reasoning, and action, offering a new framework for embodied intelligence. However, this approach requires large-scale 3D embodied datasets that are expensive to collect and thus difficult to scale. SpatialVLA \cite{qu2025spatialvla} injects Ego3D position encoding and adaptive action grids to improve spatial reasoning and trajectory generalization, but its dependence on large quantities of real-world robot trajectories makes it resource-intensive and less adaptable to new domains. PointVLA \cite{li2025pointvla}, on the other hand, augments pretrained VLAs with point cloud features through lightweight injection blocks, effectively improving geometric awareness. Yet, it relies on specialized 3D sensors such as LiDAR, which restricts training to datasets with 3D annotations and prevents large-scale pretraining with the abundant 2D corpora that have driven the success of VLAs. Collectively, these methods highlight the importance of 3D awareness but also reveal significant barriers: reliance on expensive data collection, limited scalability, or difficulty in fully leveraging existing 2D datasets.

In this paper, we address these limitations by introducing AugVLA-3D, a novel framework that enhances VLA models with sensor-free 3D structural features while preserving their compatibility with large-scale 2D training data. Specifically, our approach leverages a state-of-the-art depth estimation model (VGGT) to transform 2D RGB inputs into dense 3D point clouds, which are subsequently encoded by a PointNet-based extractor to yield compact geometric descriptors. This design bypasses the need for specialized 3D hardware and enables maximal reuse of existing 2D VLA datasets at scale. To ensure that the extracted 3D features are task-aligned and do not interfere with pretrained representations, we introduce an Action Assistant module that mirrors the main action head but with fewer parameters. Acting as a lightweight regularizer, it constrains the 3D features through structural alignment and layer-wise feature injection, thereby improving stability and downstream performance. By combining sensor-free 3D feature extraction with auxiliary regularization, AugVLA-3D achieves tight integration of semantic and geometric reasoning, leading to substantially improved generalization in real-world 3D environments. The comparison of our model with other models is shown in Fig.~\hyperref[figure:figure1]{1}. Our paper presents several notable contributions, which are as follows:

\begin{enumerate}
\item We propose a sensor-free 3D feature extraction method that leverages a depth estimation model (VGGT) to convert 2D RGB images into point clouds, from which compact 3D features are derived to enhance the original VLA model.  

\item We design a new module called Action Assistant to regularize the extracted 3D features, which constrains the learned 3D representations with action priors and ensures their consistency with downstream control tasks. 
\end{enumerate}

\begin{figure*}[t]
\begin{center}
\hypertarget{figure:figure2}{}
\label{figure:figure2}
\centering
\includegraphics[width=0.85\textwidth]{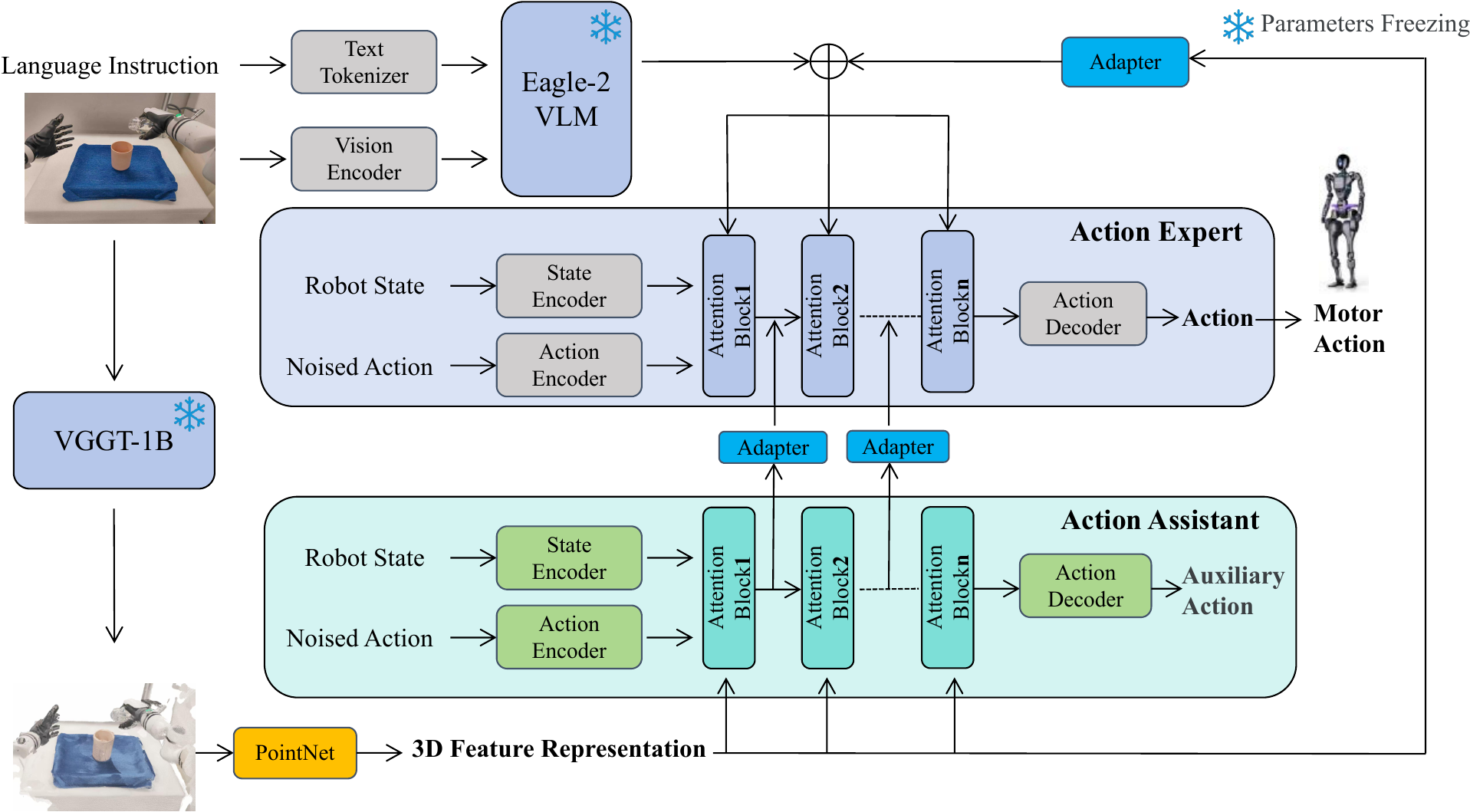}
\end{center}
\caption{Architecture of our proposed AugVLA-3D framework. The overall model design largely follows the GR00t backbone, while we introduce a dedicated 3D feature injection module to enhance the Action Expert with depth-derived geometric information. To ensure that the injected 3D features are aligned with task objectives without introducing excessive computational overhead, we further design an Action Assistant. This module is structurally consistent with the Action Expert but adopts a lightweight parameterization, effectively constraining the 3D features while keeping the additional cost minimal.}
\end{figure*}

\section{Related Works}
\textbf{2D Vision-Language-Action (VLA) Models.}
In recent years, researchers have increasingly explored training general-purpose robotic control policies from large-scale robot learning datasets \cite{dasari2019robonet,brohan2022rt,zitkovich2023rt,walke2023bridgedata,zhou2023train}. Vision-Language-Action (VLA) models \cite{yue2024deer,chiang2024mobility,zhao2025cot,lu2025vla,sautenkov2025uav,fan2025interleave}, extending visual-language models (VLMs), have emerged as a promising approach by leveraging internet-scale data and fine-tuning on robot manipulation datasets, demonstrating strong generalization and adaptability. OpenVLA \cite{kim2024openvla}, trained on 4,000 hours of open-source data, established an early baseline for generalist VLAs. More advanced methods such as $\pi$0 \cite{black2410pi0} and $\pi$0.5 \cite{black2025pi0} scale up to 10,000 hours of internal data to further improve grounding and generalization, but remain computationally heavy and reliant on large-scale pre-training. To improve efficiency, TinyVLA \cite{wen2025tinyvla} eliminates pre-training while achieving faster inference and comparable performance, and SmolVLA \cite{shukor2025smolvla} accelerates inference by skipping selected VLM layers. Despite these advances, current approaches are fundamentally built on 2D VLMs, lacking explicit 3D structural reasoning and thus struggling to ensure robust generalization in real-world environments. This limitation has motivated a shift toward integrating richer spatial awareness into VLA frameworks.

\textbf{3D-Enhanced VLA Models.}
Developing resilient visuomotor policies in 3D settings \cite{huang2023embodied,goyal2023rvt,chen2024sugar} has emerged as a core challenge in robotic learning. Recent studies increasingly strive to integrate perception, reasoning, and action into unified 3D-aware models. A representative example is 3DVLA \cite{zhen20243d}, which builds a holistic vision-language-action framework capable of tackling generalization, visual question answering, scene interpretation, and robot control in a single system. Building on this trend, iDP3 \cite{ze2024generalizable} reinforces the 3D vision backbone and demonstrates strong adaptability on humanoid robots across both egocentric and third-person viewpoints. SpatialVLA \cite{qu2025spatialvla} further incorporates Ego3D positional encodings and adaptive action grids to promote spatial reasoning, though its dependence on large collections of real-world trajectories limits scalability. PointVLA \cite{li2025pointvla} instead fuses point-cloud information into pretrained VLAs via lightweight adapters, but the need for dedicated 3D sensors and the inability to reuse massive 2D datasets remain key bottlenecks. Overall, while 3D-enhanced VLAs move closer to bridging the gap between perception and action in realistic environments, their reliance on specialized sensors and heavy data collection hinders broader applicability.

\textbf{VLA Models for Dexterous Hands.}
Dexterous manipulation \cite{yang2024task,huang2025fungrasp,zhang2025catch,bjorck2025gr00t} is a key challenge in robotics, requiring precise control and generalization across objects and tasks. GR00T-N1 \cite{bjorck2025gr00t} is a Vision-Language-Action foundation model for humanoid robots that jointly interprets instructions and generates real-time motor actions, trained on a mix of real-robot, human video, and synthetic data. Being-H0 \cite{luo2025being} treats human hands as the ultimate “foundation manipulators,” learning hand motions from large-scale human videos and transferring these skills to robots. DexVLG \cite{he2025dexvlg} constructs a massive dataset of 170 million grasp poses mapped to 174,000 simulated objects with detailed part-level descriptions, enabling instruction-aligned dexterous grasping. These works collectively demonstrate the potential of data-driven and foundation-model approaches for advancing general-purpose dexterous manipulation. More broadly, they also highlight a recurring theme across the VLA literature: the need to balance scalability, efficiency, and embodiment-specific adaptability, which continues to drive the design of next-generation robotic foundation models.

\section{Methods}
A key limitation of existing Vision-Language-Action (VLA) models lies in their reliance on 2D visual features, which lack explicit 3D structural reasoning and thus hinder generalization in real-world environments. To address this, we propose a novel 3D Feature Injection framework that augments VLA models with depth-based geometric cues while maintaining compatibility with large-scale 2D training corpora. Specifically, we leverage a state-of-the-art depth estimation model (VGGT) to transform standard 2D RGB observations into dense 3D point clouds, enabling sensor-free acquisition of structural information. A PointNet encoder is then employed to extract compact 3D features, which serve as geometric priors for action prediction. To further align these 3D features with task objectives, we introduce a new module called Action Assistant that mirrors the Action Expert head but with fewer parameters. Acting as a lightweight regularizer, this module constrains the learned 3D features and outputs intermediate activations that are injected, together with PointNet features, into the corresponding layers of the original VLA backbone. This design achieves a tight integration of 2D semantics and 3D geometry, reinforcing multimodal fusion while avoiding dependence on 3D sensors. The architecture of our model is shown in Fig.~\hyperref[figure:figure1]{2}.

\subsection{3D Feature Extraction based on Depth estimation}
Our framework is motivated by a critical limitation of current vision-language-action (VLA) models: their reliance on 2D features inherited from large-scale vision-language models (VLMs). While such features provide strong semantic grounding, they lack explicit geometric reasoning, which is essential for accurate interaction in 3D environments. As a result, current VLA models often struggle in tasks requiring precise depth understanding, collision avoidance, or spatial planning. For instance, ambiguous 2D cues can lead to failures when objects overlap in the image plane or when perspective distortions alter apparent distances, highlighting the need for explicit structural priors.

Recent attempts have begun addressing this gap by incorporating 3D information into VLA architectures. Among them, PointVLA augments pretrained VLAs with LiDAR-derived point clouds through lightweight injection blocks. Although this approach improves geometric awareness, it introduces substantial practical limitations. Specifically, reliance on LiDAR sensors restricts applicability to domains where high-quality 3D data is available, and the resulting datasets are typically limited in scale. Consequently, such methods can only be applied during fine-tuning, preventing the full reuse of the massive 2D datasets that have driven recent progress in VLMs and VLAs.

To overcome these shortcomings, we propose a sensor-free approach to infuse 3D reasoning into VLA models. Our framework leverages VGGT, a state-of-the-art monocular depth estimator, to predict dense depth maps from standard RGB images. The estimated depth maps are back-projected into camera-centered point clouds using known intrinsics, followed by light preprocessing steps such as outlier filtering and normalization. These point clouds are then encoded by a PointNet backbone, producing compact descriptors that capture both local geometric structures and global spatial layouts. By converting ubiquitous 2D data into 3D representations, our method enables large-scale reuse of existing VLA training corpora while providing explicit structural cues absent in purely 2D models. The extracted 3D feature $f_{3D}$ is formulated as follows:

\begin{equation}
\begin{aligned}
    P &= f\big((I_i)_{i=1}^N\big), \quad && I_i \in \mathbb{R}^{3 \times H \times W} \\
    \tilde{P} &= \mathcal{S}(P), \quad && \tilde{P} \in \mathbb{R}^{M' \times 3}, \; M' < H \times W \\
    f_{3D} &= \text{PointNet}(\tilde{P}), \quad && F_{3D} \in \mathbb{R}^{M' \times C}
\end{aligned}
\end{equation}

where $f(\cdot)$ denotes the VGGT depth estimation model, $(I_i)_{i=1}^N$ represents a sequence of $N$ RGB images and $N$ represents the number of observation perspectives of the robot; if there is only one observation perspective, $N$ equals 1. $P$ is the raw point cloud map obtained from VGGT, $\mathcal{S}(\cdot)$ is a point cloud sampling operator used to reduce the number of points to $M'$ and $C$ is the dimension encoded by PointNet.

Our framework achieves a tight integration of 2D semantics and 3D structure without requiring dedicated 3D sensors. By transforming RGB data into actionable geometric priors, the model gains robust spatial reasoning while retaining the scalability of 2D training pipelines. Compared to LiDAR-dependent approaches such as PointVLA, our method is more broadly applicable: it leverages monocular depth estimation to scale 3D feature injection to the full extent of existing 2D corpora. This sensor-free design maximizes scalability, reduces deployment costs, and substantially improves generalization in real-world 3D environments.

\subsection{Action Assistant}
While depth-derived features introduce valuable geometric priors, injecting them directly into a pretrained VLA backbone often leads to unstable optimization and performance degradation, as raw geometric signals are not naturally aligned with the task-specific action representation space. To mitigate this, we design a new module called the Action Assistant that mirrors the Action Expert head but with a lightweight parameterization. This auxiliary expert acts as a task-aligned regularizer: it leverages PointNet-extracted 3D features to generate motion constraints, and both the original PointNet features and the intermediate activations from each layer of the motion assistant are injected into the corresponding layers of the primary motion expert. This design constrains the learning of 3D features, ensuring that they align with manipulation objectives and enhance the pretrained 2D pathway rather than disrupting it.

The action assistant mirrors the structure of the primary action head but uses significantly fewer parameters. Its role is twofold: (1) to act as a \textit{task-guided projector} that transforms PointNet-extracted 3D features into action-relevant embeddings, and (2) to serve as an \textit{intermediate regularizer} that injects stable guidance into the VLA backbone at multiple depths. Concretely, for the $l$-th layer, we compute

\begin{equation}
\tilde{h}^{(l)} = h_{\text{orig}}^{(l)} + \alpha^{(l)} \cdot \mathcal{T}\big(h_{\text{aux}}^{(l)}, f_{\text{3D}}\big),
\end{equation}

where $h_{\text{orig}}^{(l)}$ is the original hidden state of the VLA, $h_{\text{aux}}^{(l)}$ is the corresponding activation from the auxiliary expert, $f_{\text{3D}}$ denotes the PointNet features, and $\alpha^{(l)}$ is a learnable scalar gate. The transformation $\mathcal{T}(\cdot)$ is implemented as a lightweight projection or cross-attention module, ensuring that injected features are smoothly aligned with the pretrained representation.
\begin{figure*}[t]
\label{figure:figure3}
\centering
\includegraphics[width=0.9\textwidth]{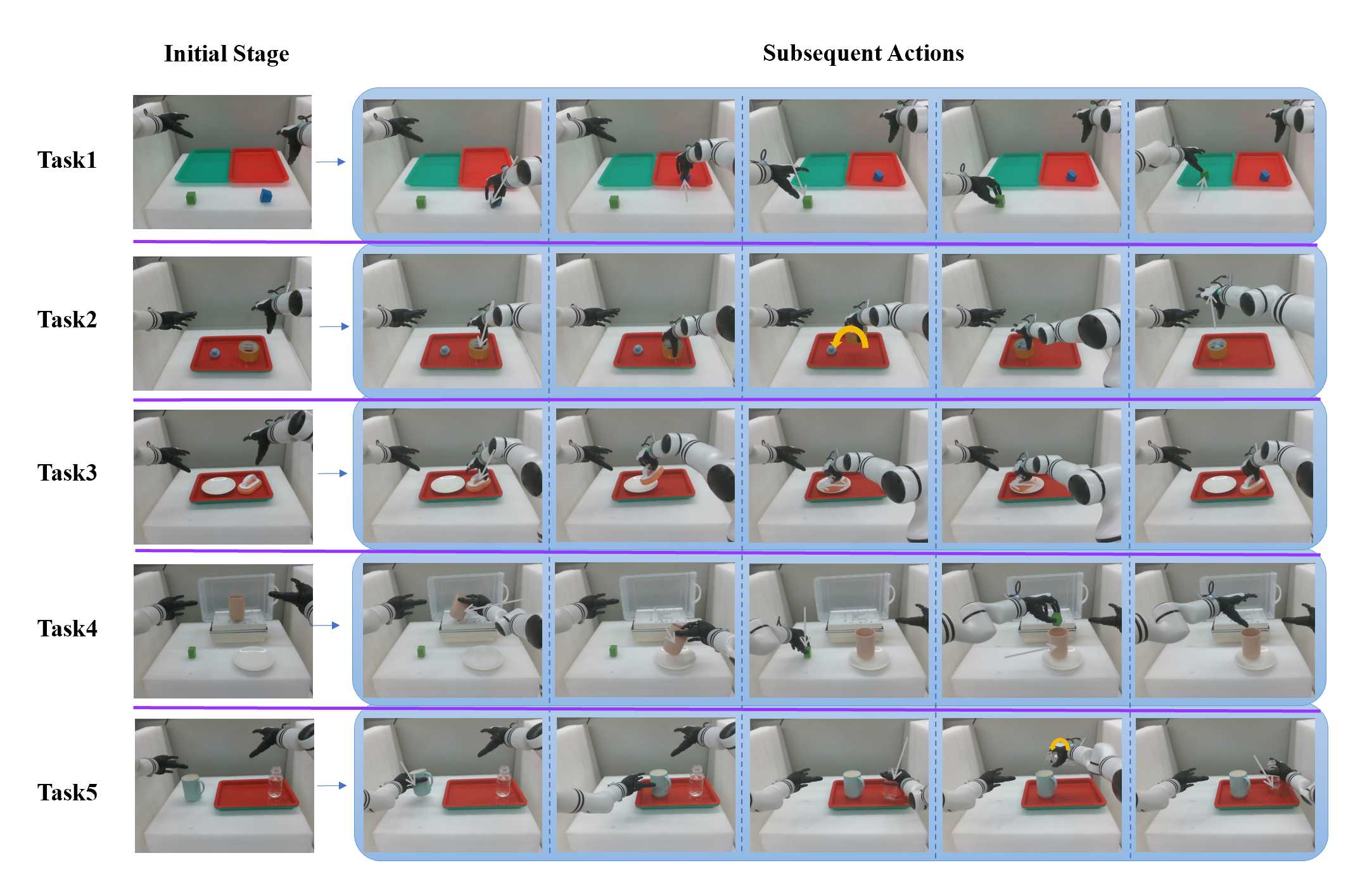}
\caption{Illustrations of the five experimental tasks: 
Task 1: Place the wooden blocks into the corresponding plates; 
Task 2: Cover the duck toy with a tape; 
Task 3: Wipe the plates with a dishcloth; 
Task 4: Take out the cup and put in the block; 
Task 5: Place the cup and pour water into it.}
\label{fig:tasks}
\end{figure*}

In order to minimize the additional parameter and computational cost, the action assistant adopts a compact \textit{transformer-diffusion} architecture. Specifically, we reduce the hidden dimensions, share weights across denoising steps, and limit the diffusion horizon to a few iterations. This structure not only maintains efficiency but also leverages the generative prior of diffusion modeling to provide stronger regularization of 3D-to-action mappings.

Critically, the auxiliary actions generated by the action assistant are used solely for computing auxiliary losses to constrain 3D feature learning, and these actions are never directly executed or used to update the motor commands of the robot. This design ensures that the primary action expert retains full control over policy execution while still benefiting from the regularization provided by the assistant. The separation preserves the stability of the pretrained VLA backbone, as the outputs from the Assistant influence feature representations without bypassing the decision-making process of the main policy.

\begin{table*}
    \centering
    \caption{Simulation results comparing our method (AugVLA-3D) with Diffusion Policy and Gr00T across various manipulation tasks. AugVLA-3D, enhanced with 3D features, outperforms Gr00T in most tasks, demonstrating the effectiveness of our approach.}
    \label{table:table1}
    \begin{tabular}{c|cc|cc|cc}
        \multirow{2}{*}{Methods}    & \multicolumn{2}{c}{Diffusion Policy} & \multicolumn{2}{c}{Gr00T}  & \multicolumn{2}{c}{AugVLA-3D}\\ \\[-0.2cm] \cline{2-7} \\[-0.2cm]  \\[-0.2cm]
            & \multicolumn{1}{c}{30 demos}& \multicolumn{1}{c|}{100 demos} & \multicolumn{1}{c}{30 demos}& \multicolumn{1}{c|}{100 demos}& \multicolumn{1}{c}{30 demos}& \multicolumn{1}{c}{100 demos}\\ \\[-0.2cm] \hline \\[-0.2cm]
\multirow{1}{*}Cutting Board to Pot              & 23\% & 37\%  & \cellcolor{gray!20}59\% & 58\%  & 57\%  & \cellcolor{gray!20}60\% \\
\multirow{1}{*}Cutting Board to Basket         &20\%  & 42\%  & 43\% & 62\%  & \cellcolor{gray!20}46\%  & \cellcolor{gray!20}65\% \\
\multirow{1}{*}Cutting Board to Tiered Basket & 14\% & 14\%  & 14\% & 24\%  & \cellcolor{gray!20}18\%  & \cellcolor{gray!20}28\% \\
\multirow{1}{*}Cutting Board to Pan              & 28\% & 48\%  & 68\% & 66\%  & \cellcolor{gray!20}70\%  & \cellcolor{gray!20}71\% \\
\multirow{1}{*}Cutting Board to Cardboard Box &12\%  & 16\%  & 31\% & 30\%  & \cellcolor{gray!20}36\%  &\cellcolor{gray!20}33\%\\
\multirow{1}{*}Placemat to Bowl                 & 15\% & 19\%  & 31\% & 39\%  & \cellcolor{gray!20}34\%  & \cellcolor{gray!20}39\%\\    \\[-0.2cm] \hline \\[-0.2cm]
\multirow{1}{*}Placemat to Plate                & 16\% & 24\%  & 33\% & 37\%  & \cellcolor{gray!20}38\%  & \cellcolor{gray!20}41\%\\
\multirow{1}{*}Placemat to Basket               &16\%  & 24\%  & \cellcolor{gray!20}50\% & 46\%  & 49\%  & \cellcolor{gray!20}52\%\\
\multirow{1}{*}Placemat to Tiered Shelf        & 7\%  & 6\%   & 12\% & \cellcolor{gray!20}22\%  &\cellcolor{gray!20}18\%  & \cellcolor{gray!20}25\%\\
\multirow{1}{*}Plate to Pan                     &14\%  & 18\%  & 35\% & \cellcolor{gray!20}48\%  & \cellcolor{gray!20}38\%  & 47\%\\
\multirow{1}{*}Plate to Cardboard Box          & 13\% &14\%   &34\%  & 38\%  & \cellcolor{gray!20}39\%  & \cellcolor{gray!20}39\% \\
\multirow{1}{*}Plate to Bowl                   & 16\% & 19\%  & \cellcolor{gray!20}41\% & 42\%  & 40\%  & \cellcolor{gray!20}43\% \\    \\[-0.2cm] \hline \\[-0.2cm]
\multirow{1}{*}Plate to Plate                   & 26\% &39\%   & \cellcolor{gray!20}73\% & 85\%  & 72\%  & \cellcolor{gray!20}87\% \\
\multirow{1}{*}Tray to Tiered Shelf            & 2\%  & 7\%   & 18\% & 28\%  & \cellcolor{gray!20}22\%  & \cellcolor{gray!20}31\% \\
\multirow{1}{*}Tray to Tiered Basket           & 13\% & 34\%  & 33\% & 49\%  & \cellcolor{gray!20}36\%  & \cellcolor{gray!20}52\% \\
\multirow{1}{*}Tray to Plate                   & 27\% & 41\%  & 54\% & \cellcolor{gray!20}69\%  & \cellcolor{gray!20}60\%  & 67\%\\
\multirow{1}{*}Tray to Cardboard Box           & 22\% & 37\%  & 51\% & 56\%  & \cellcolor{gray!20}56\%  & \cellcolor{gray!20}60\%\\
\multirow{1}{*}Tray to Pot                     & 22\% & 48\%  & 52\% & 60\%  & \cellcolor{gray!20}55\%  & \cellcolor{gray!20}65\%\\    \\[-0.2cm] \hline \\[-0.2cm]
\multirow{1}{*}Wine to Cabinet                 & 43\% & 56\%  & 58\% & 54\%  & \cellcolor{gray!20}60\%  & \cellcolor{gray!20}62\%\\
\multirow{1}{*}Place Bottle to Cabinet        & 40\% & 63\%  & \cellcolor{gray!20}61\% & 81\%  & 60\%  & \cellcolor{gray!20}83\%\\
\multirow{1}{*}Place Milk to Microwave          &37\%  & 41\%  & 42\% & 59\%  & \cellcolor{gray!20}45\%  & \cellcolor{gray!20}63\%\\
\multirow{1}{*}Potato to Microwave             & 18\% & 30\%  &30\%  & 27\%  & \cellcolor{gray!20}35\%  & \cellcolor{gray!20}35\% \\
\multirow{1}{*}Cup to Drawer                   & 25\% & 32\%  &\cellcolor{gray!20}36\%  & 44\%  & 34\%  & \cellcolor{gray!20}46\% \\
\multirow{1}{*}Can to Drawer                   & 48\% & 75\%  &\cellcolor{gray!20}78\%  & 77\%  & 75\%  & \cellcolor{gray!20}81\% \\    \\[-0.2cm] \hline \\[-0.1cm]
\multirow{1}{*}Average                         & 21\% & 33\%  &43\%  &50\%    & \cellcolor{gray!20}46\%  & \cellcolor{gray!20}54\% \\ [+0.12cm] \hline
    \end{tabular}
\end{table*}

\subsection{Intropy Analysis of AugVLA-3D}

Under the Intropy framework~\cite{Yu2026Intropy,Ren2025Intelligence}, AugVLA-3D increases intelligence gain
$\mathrm{d}L=\delta S/R$ by enriching geometric discrepancy signals while reducing spatial
ambiguity. Depth-derived 3D features inject dense, task-relevant information into the action
pathway, increasing $\delta S$ beyond 2D vision-language alignment alone. Simultaneously,
geometry-aware feature routing and action-assisted regularization reduce effective resistance $R$ by constraining optimization with physically meaningful structure. The result is higher Intropy, yielding more robust and generalizable 3D manipulation.

\section{Experiments}
This section evaluates the effectiveness of our approach through a series of experiments, including real-world physical trials across diverse scenarios and multi-task validation in the RoboCasa simulation environment, following the experimental protocol of Gr00T \cite{bjorck2025gr00t}.

\subsection{Implementation Details}
Our model is built upon the GR00t backbone, with the proposed 3D feature injection module integrated into its action prediction pipeline. All experiments are conducted on a single NVIDIA RTX 4090 GPU. Due to limited computational resources, we train our framework using only the \texttt{PhysicalAI-Robotics-GR00T-X-Embodiment-Sim} dataset, and specifically restrict training to 10\% of its data for only 1 epoch. While this constraint prevents us from scaling training to larger and more diverse datasets, or from conducting extensive evaluations across multiple benchmarks, our results nevertheless highlight the effectiveness of the proposed design. 
% We believe that training on larger-scale datasets and broader benchmarks would further amplify the advantages of our approach, which we leave for future work.

\subsection{Experimental Results on Real-Life Scenarios}
We evaluate our approach on the dexterous robotic hand ROH-A001 across five distinct tasks: Place the wooden blocks into the corresponding plates, Cover the duck toy with a tape, Wipe the plates with a dishcloth, Take out the cup and put in the block and Place the cup and pour water into it, as illustrated in Fig.~\hyperref[figure:figure3]{3}.

\textbf{ROH-A001.} The robotic hand is equipped with 11 joints and 6 active degrees of freedom, enabling fine-grained motion control and grasping capabilities comparable to those of a human hand.

\textbf{Place the wooden blocks into the corresponding plates.}
The robot picks up green and red blocks and places them into their corresponding color-marked areas. The challenge lies in identifying the right locations and executing precise placement in a 3D space, which requires dynamic coordination between visual recognition and hand positioning.

\textbf{Cover the duck toy with a tape.}
In this task, the robot grasps a roll of tape and places it over a toy duck, securing it in place. The main difficulty lies in positioning both the tape and the duck in 3D space. To align the tape correctly and ensure it covers the duck efficiently, 3D spatial awareness is essential for precise manipulation and placement.

\begin{figure}
\label{figure:figure4}
\centering
\includegraphics[width=8.5cm]{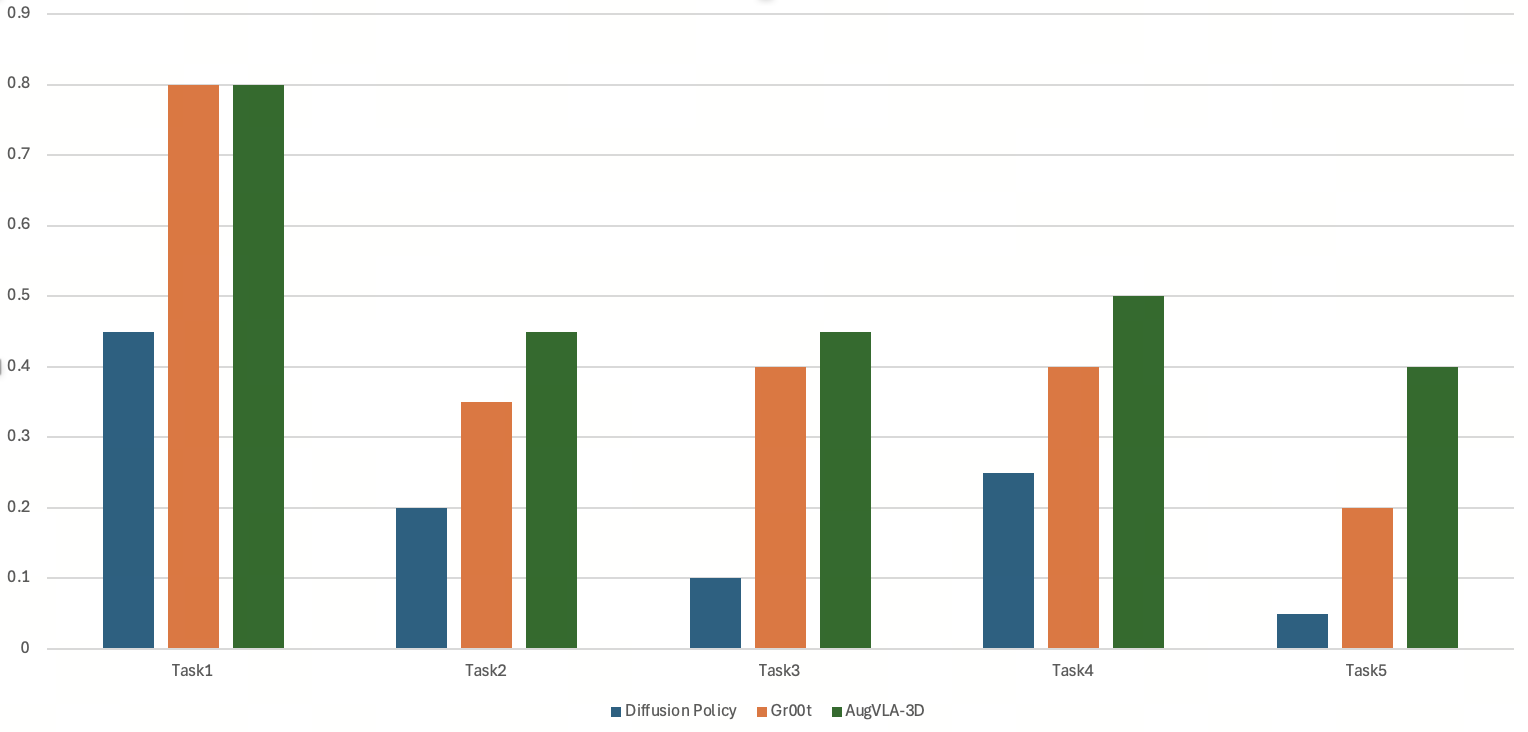}
\caption{Experimental results on real-life scenarios}
\end{figure}

\textbf{Wipe the plates with a dishcloth.}
The robot grabs a dishcloth and uses it to clean a plate. The challenge here is to account for the dishcloth’s deformation as it interacts with the plate’s surface. Using 3D features helps the robot adjust its actions based on the cloth's shape and movement, leading to a more accurate and effective cleaning process.

\textbf{Take out the cup and put in the block.}
In this task, the robot retrieves a cup from the cabinet and places a blue block inside it. The challenge is handling the precise placement of the block inside the cup, considering the cup’s 3D orientation and position, and managing any potential block collisions or tipping.

\textbf{Place the cup and pour water into it.}
The robot places a cup and pours water into it. The task requires careful control of the pouring angle and speed, while maintaining the correct position of the cup. 3D feedback helps the robot make real-time adjustments to avoid spillage and achieve a controlled pour.

\begin{figure*}[h]
    \label{figure:figure5}
    \centering
    
    % 第一组图片
    \begin{minipage}{.15\textwidth}
        \centering
        \includegraphics[width=\textwidth]{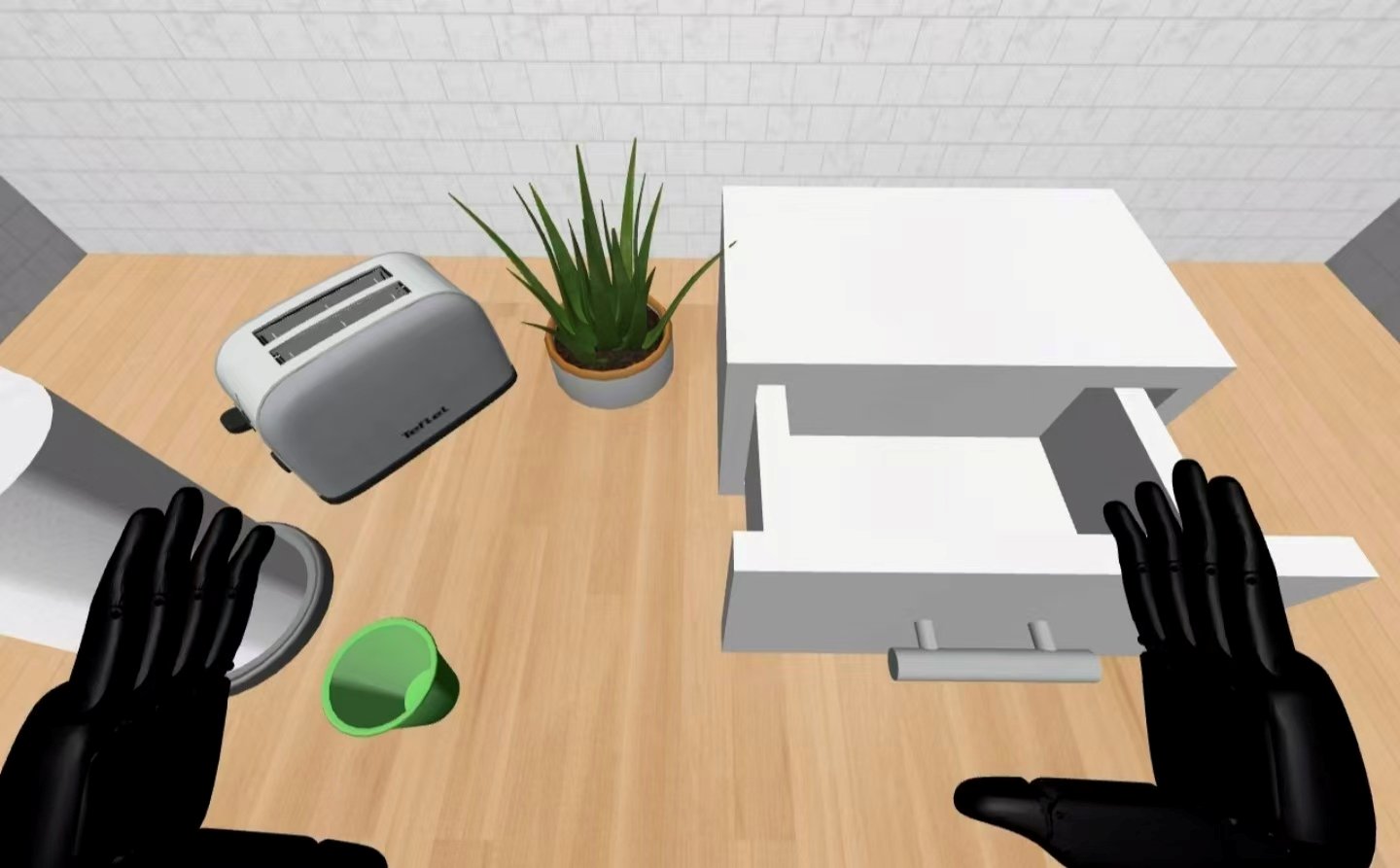}
    \end{minipage}
\hfill   
\begin{minipage}{.15\textwidth}
        \centering
        \includegraphics[width=\textwidth]{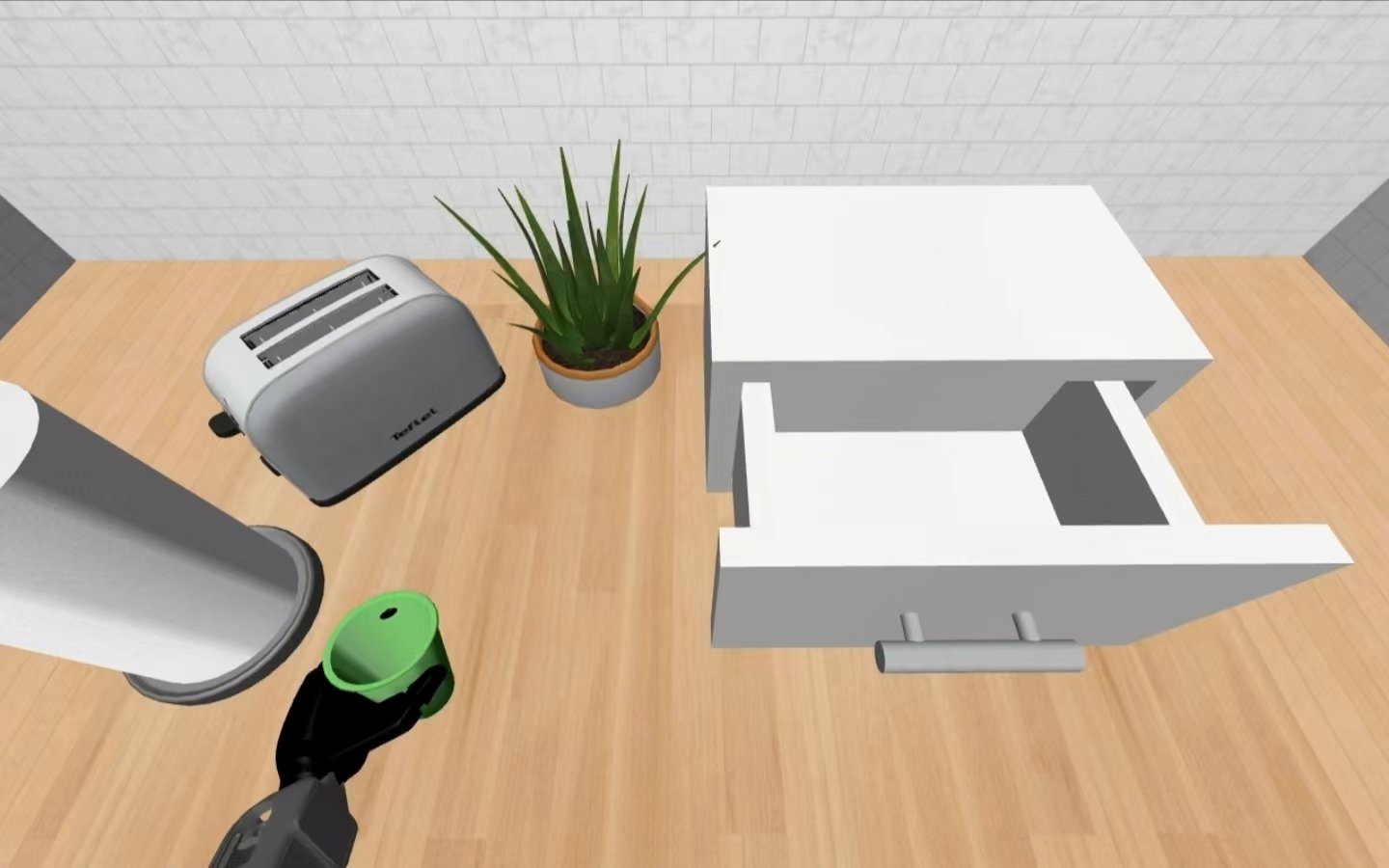}
    \end{minipage}
\hfill  
\begin{minipage}{.15\textwidth}
        \centering
        \includegraphics[width=\textwidth]{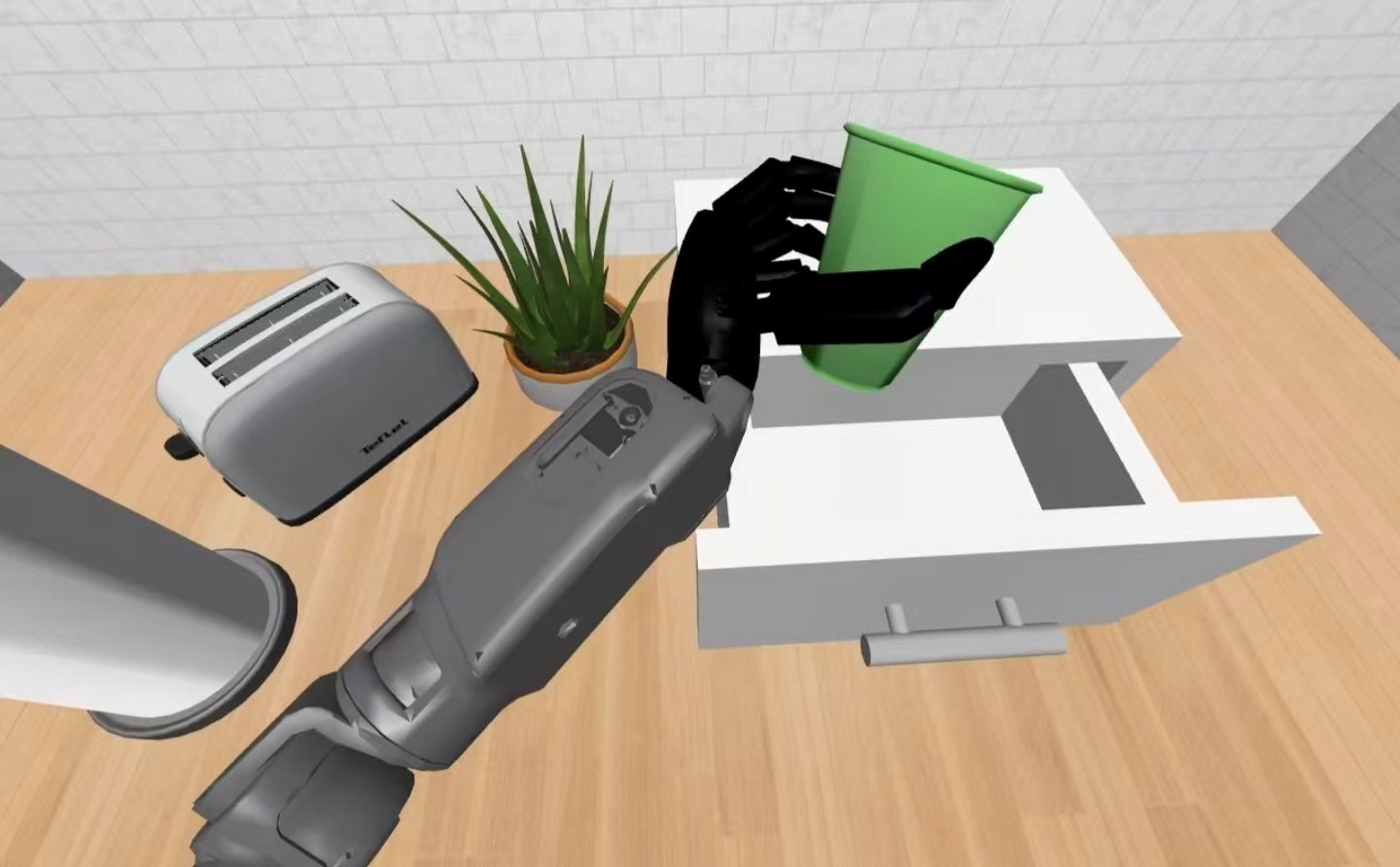}
    \end{minipage}
\hfill 
\begin{minipage}{.15\textwidth}
        \centering
        \includegraphics[width=\textwidth]{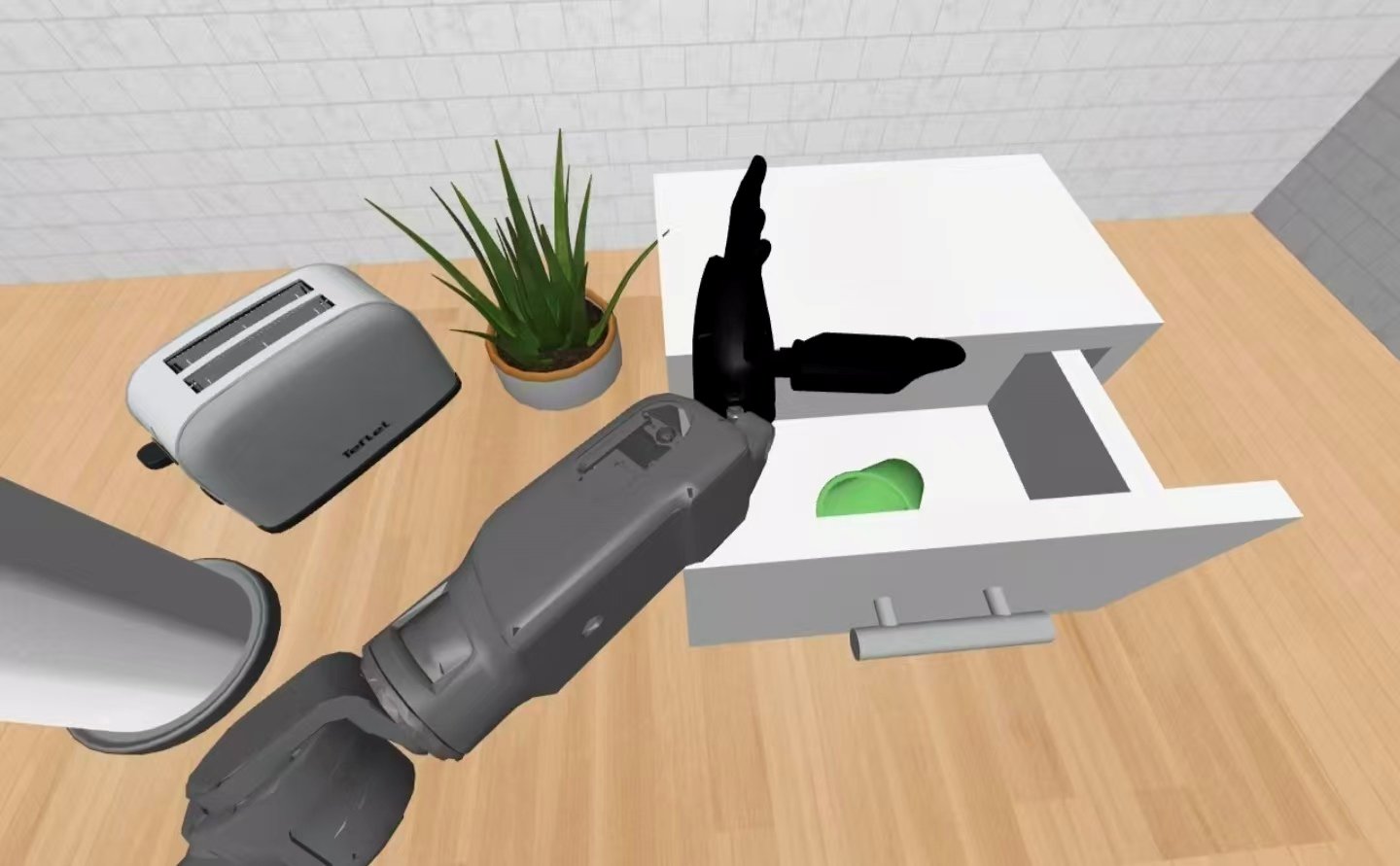}
    \end{minipage}
\hfill  
\begin{minipage}{.15\textwidth}
        \centering
        \includegraphics[width=\textwidth]{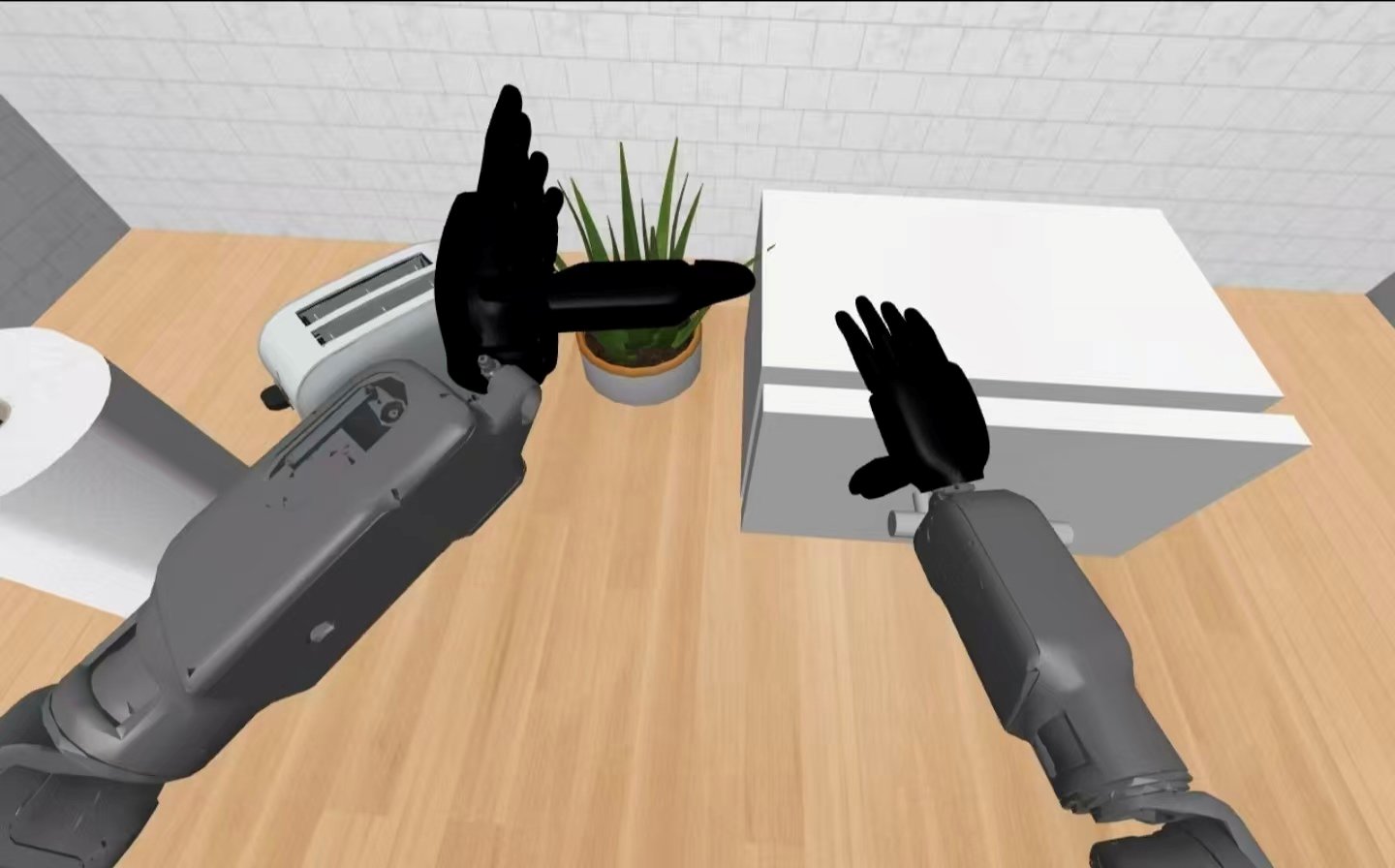}
    \end{minipage}
\hfill  
\begin{minipage}{.15\textwidth}
        \centering
        \includegraphics[width=\textwidth]{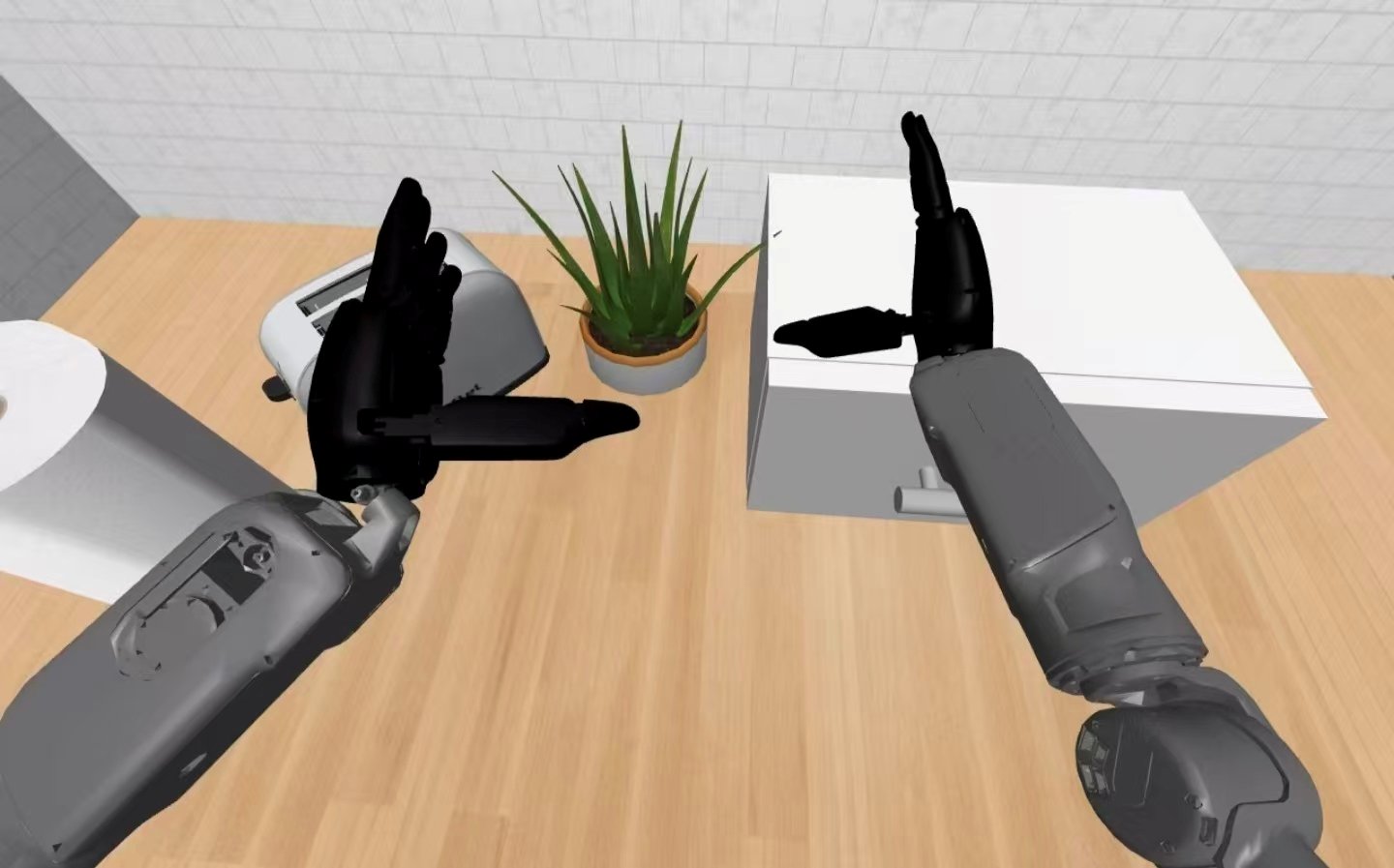}
    \end{minipage}
    
    % 第一组文字说明（左对齐）
    \raggedright % 行题注左对齐
    \par\vspace{1em}
    AugVLA-3D: pick up the cup and place it into the open drawer, then close it.
    \par\vspace{1em}
    \centering % 恢复居中对齐后续内容
    
    % 第二组图片
    
\begin{minipage}{.15\textwidth}
        \centering
        \includegraphics[width=\textwidth]{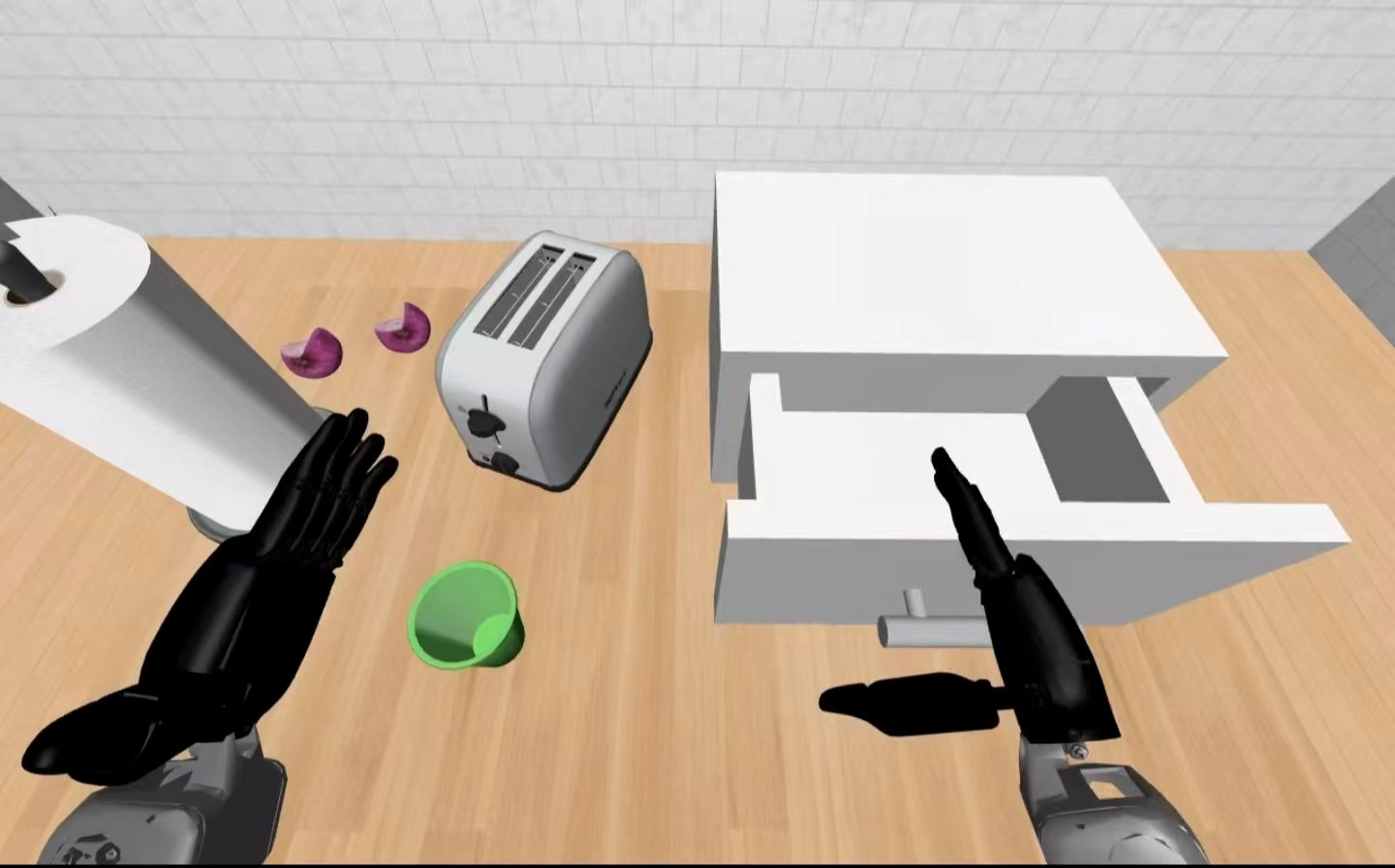}
    \end{minipage}
\hfill   
\begin{minipage}{.15\textwidth}
        \centering
        \includegraphics[width=\textwidth]{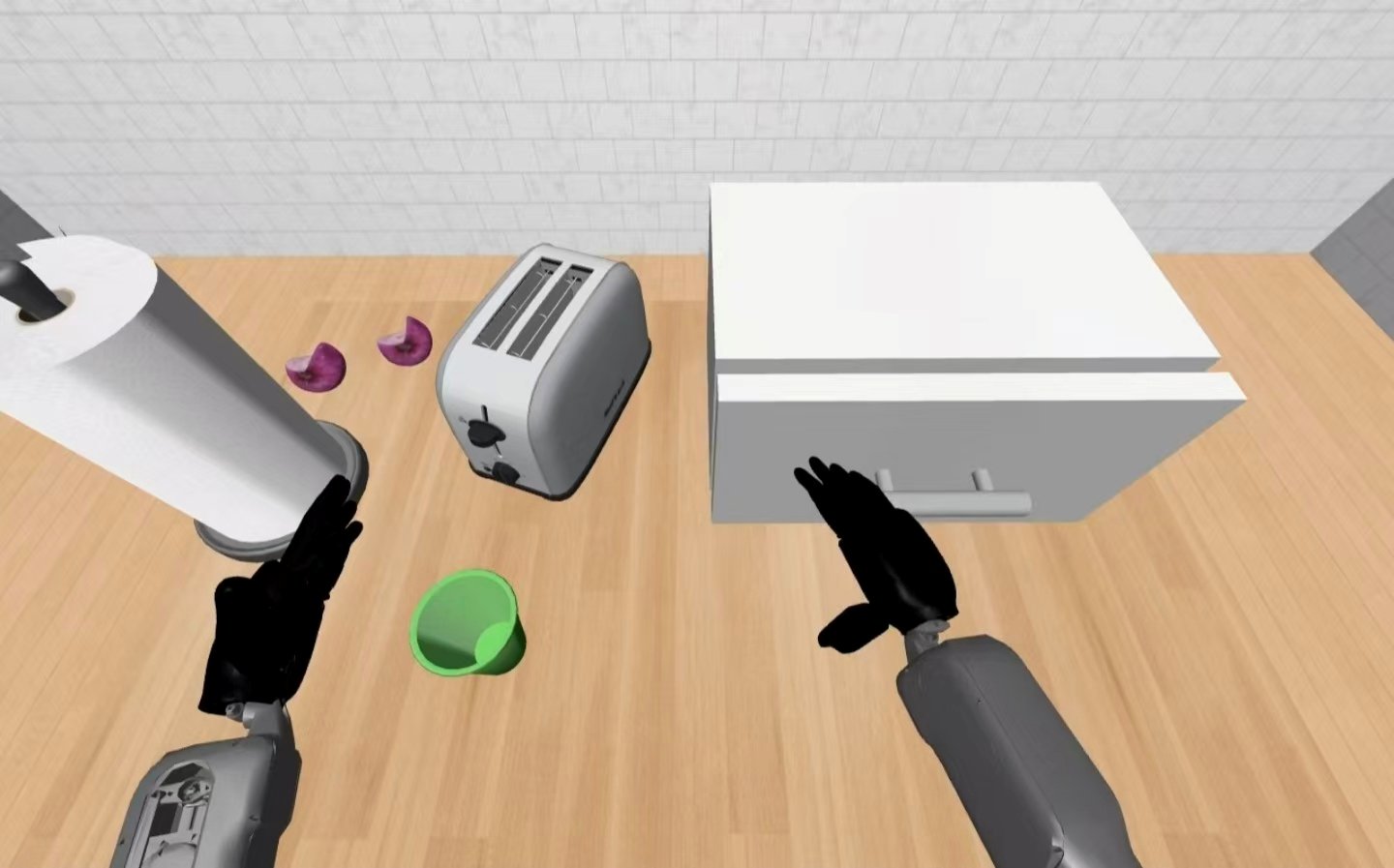}
    \end{minipage}
\hfill 
\begin{minipage}{.15\textwidth}
        \centering
        \includegraphics[width=\textwidth]{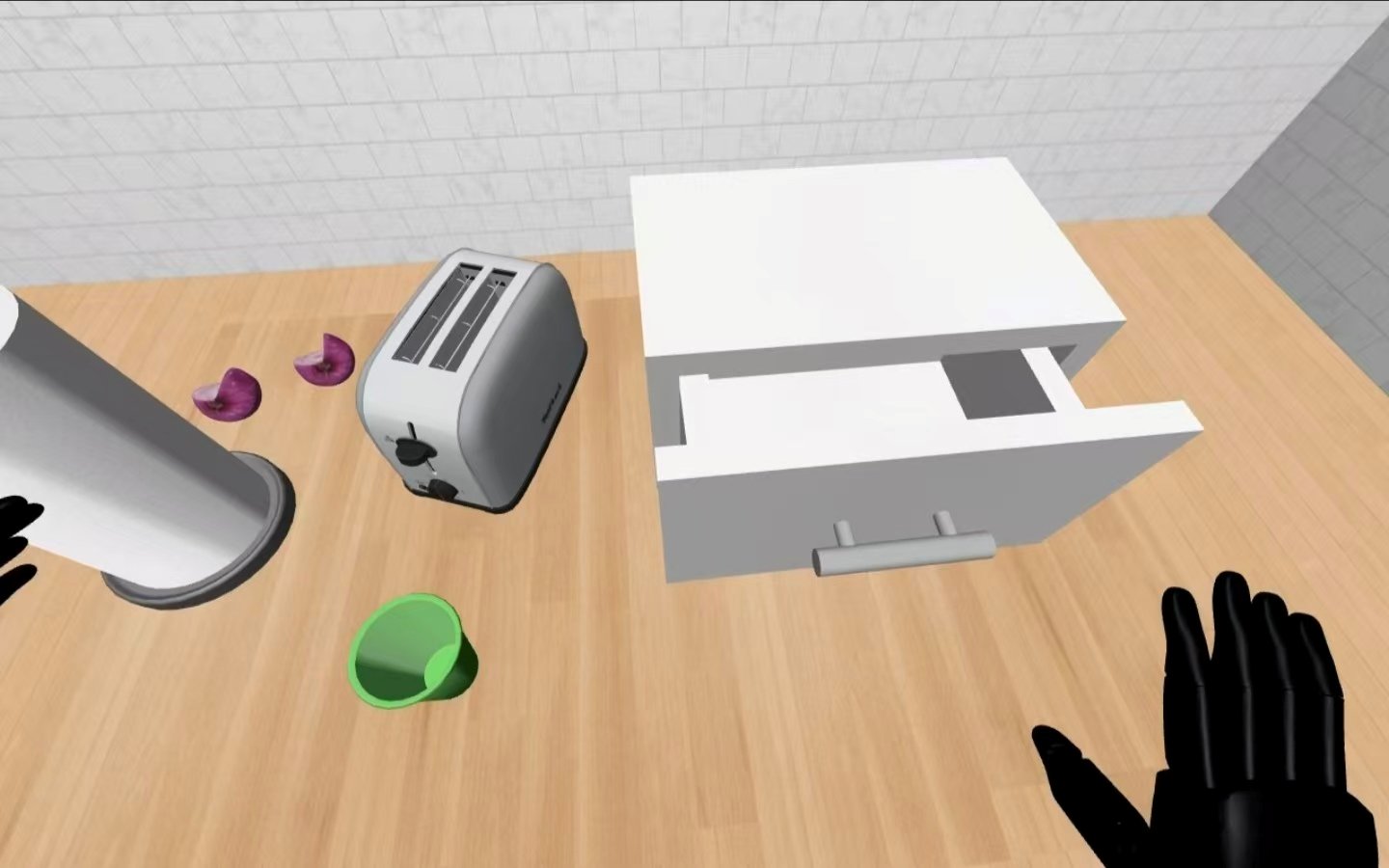}
    \end{minipage}
\hfill 
\begin{minipage}{.15\textwidth}
        \centering
        \includegraphics[width=\textwidth]{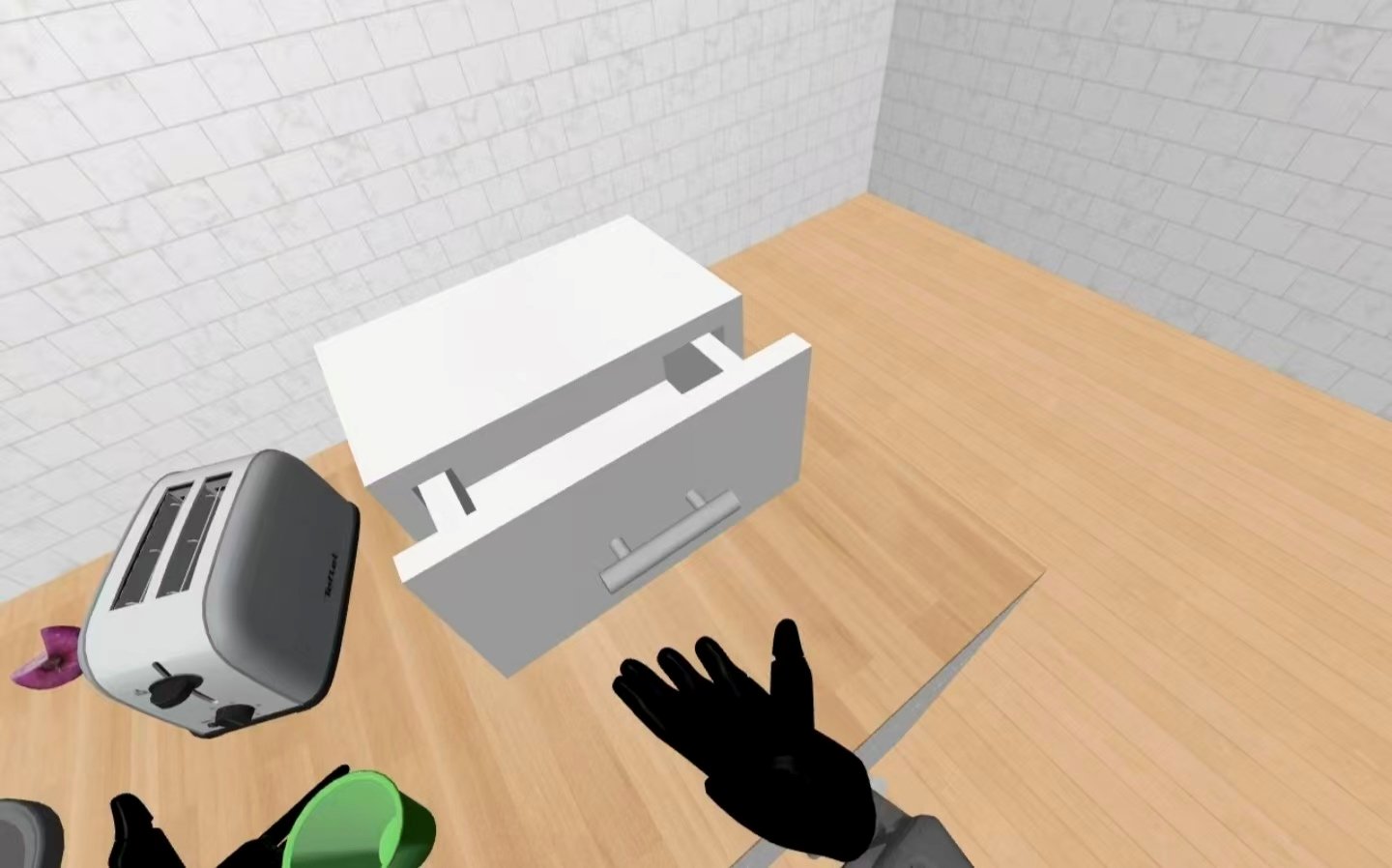}
    \end{minipage}
\hfill   
\begin{minipage}{.15\textwidth}
        \centering
        \includegraphics[width=\textwidth]{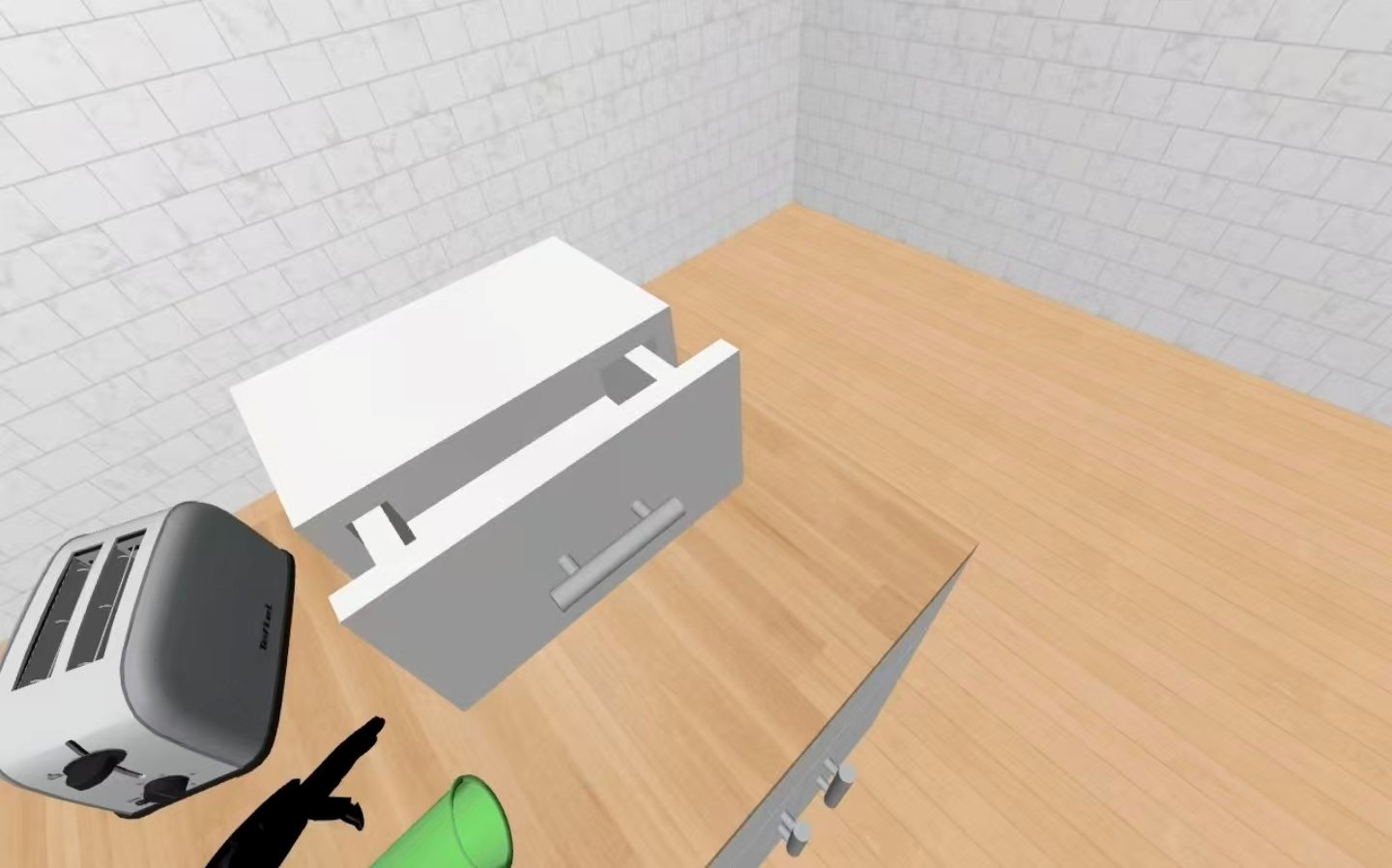}
    \end{minipage}
\hfill
\begin{minipage}{.15\textwidth}
        \centering
        \includegraphics[width=\textwidth]{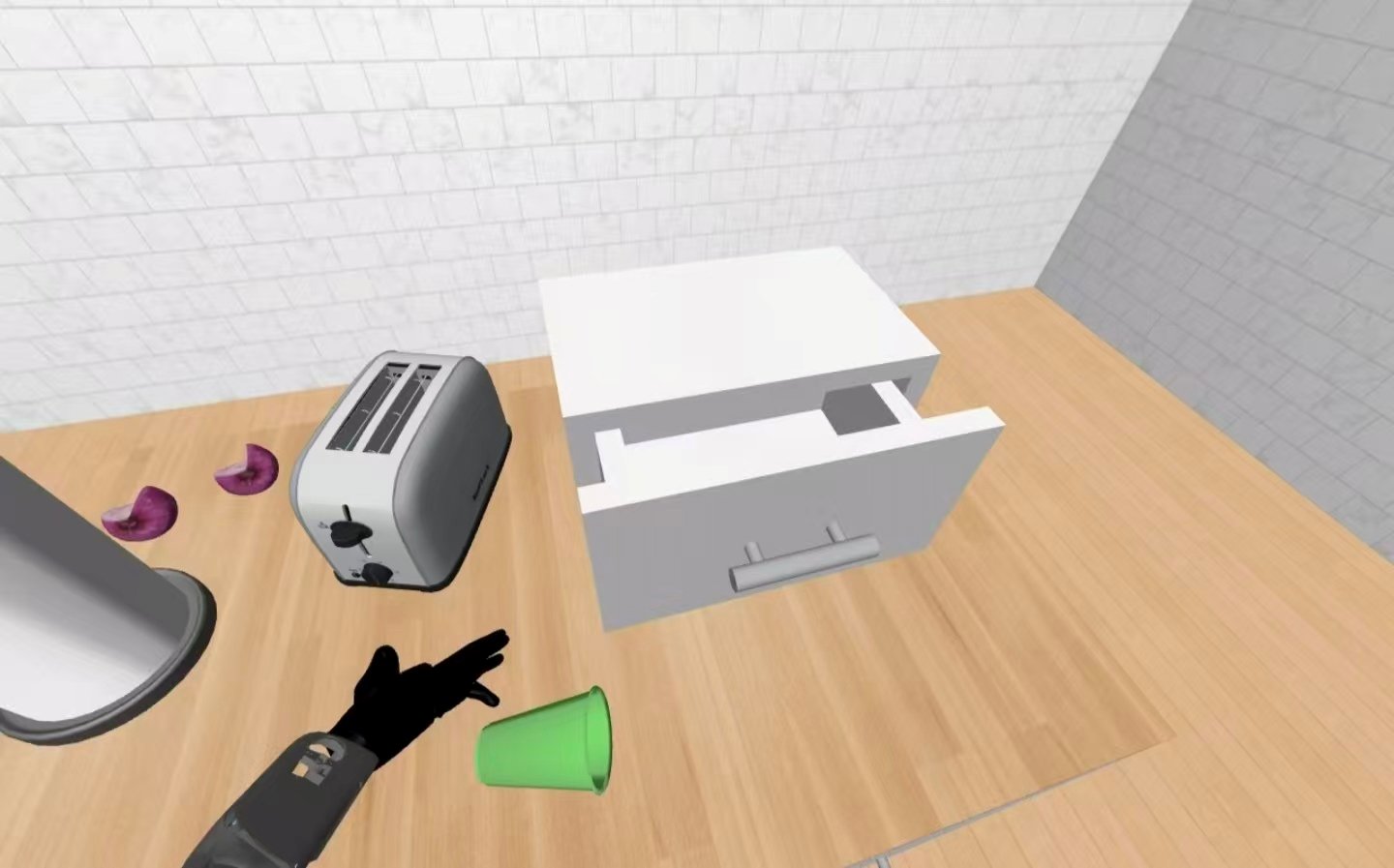}
    \end{minipage}
    
    % 第二组文字说明（左对齐）
    \raggedright
    \par\vspace{1em}
    Gr00T: pick up the cup and place it into the drawer,  then close it
    \par\vspace{1em}
    \centering
    
    % 第三组图片
    
\begin{minipage}{.15\textwidth}
        \centering
        \includegraphics[width=\textwidth]{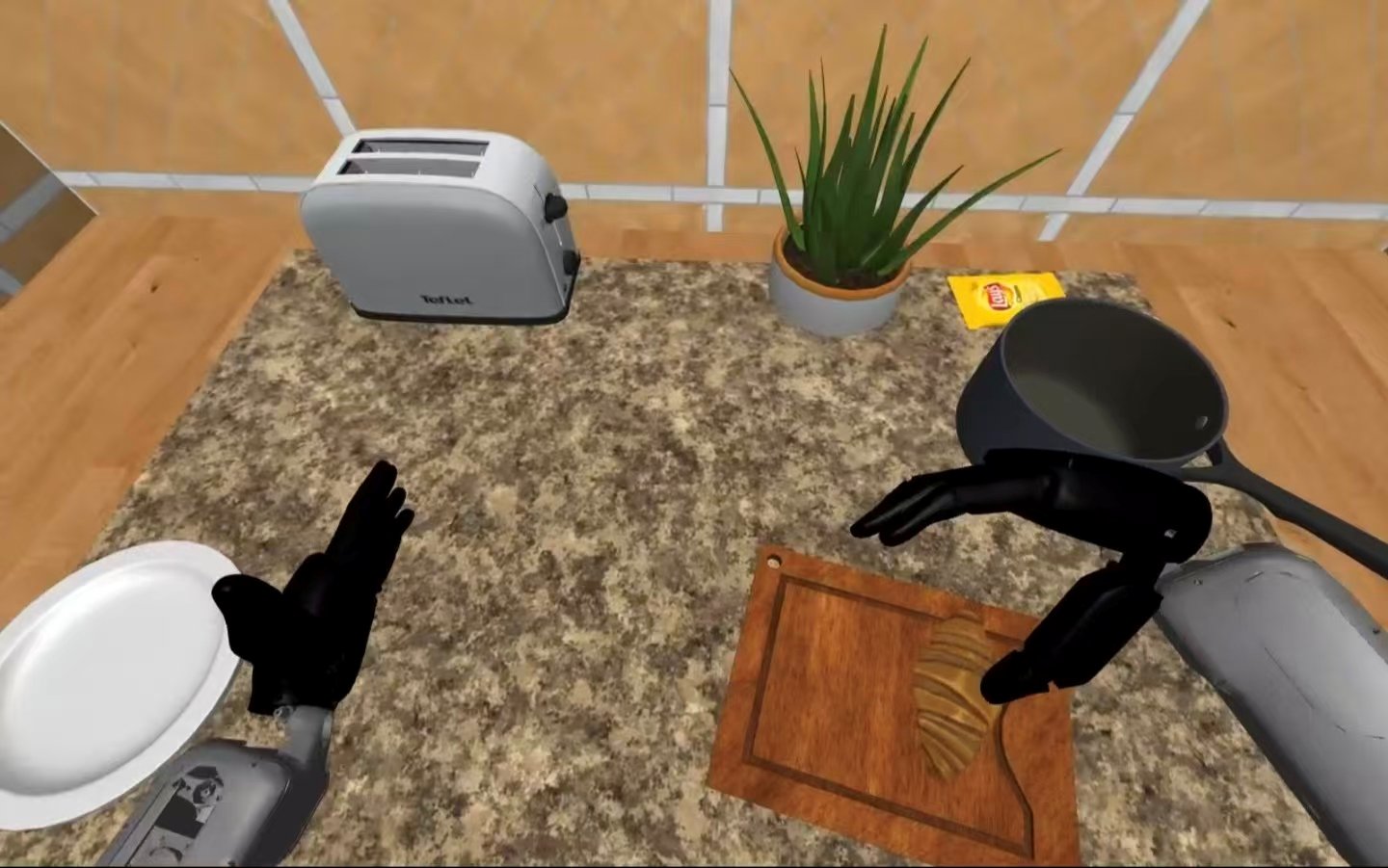}
    \end{minipage}
\hfill 
\begin{minipage}{.15\textwidth}
        \centering
        \includegraphics[width=\textwidth]{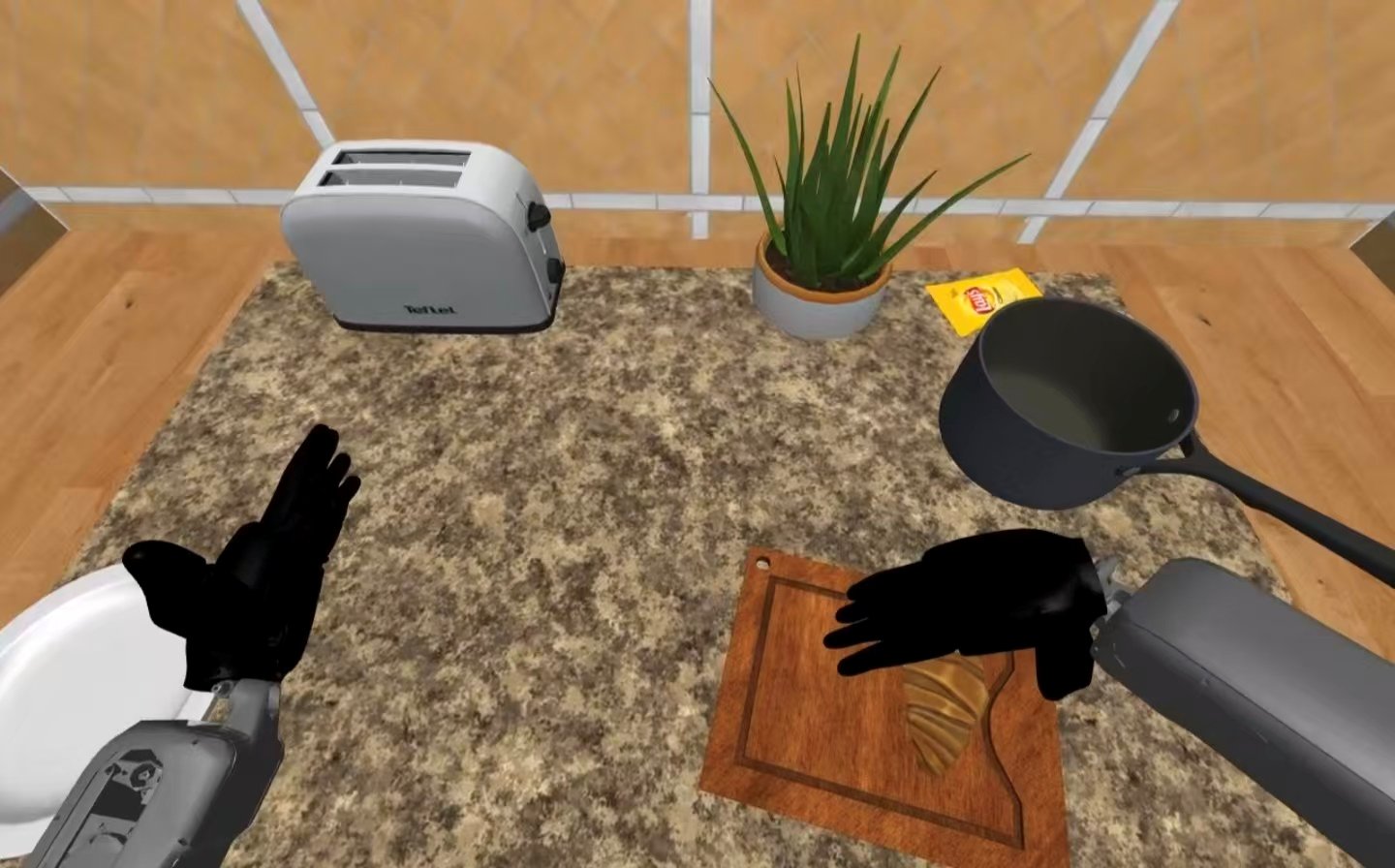}
    \end{minipage}
\hfill 
\begin{minipage}{.15\textwidth}
        \centering
        \includegraphics[width=\textwidth]{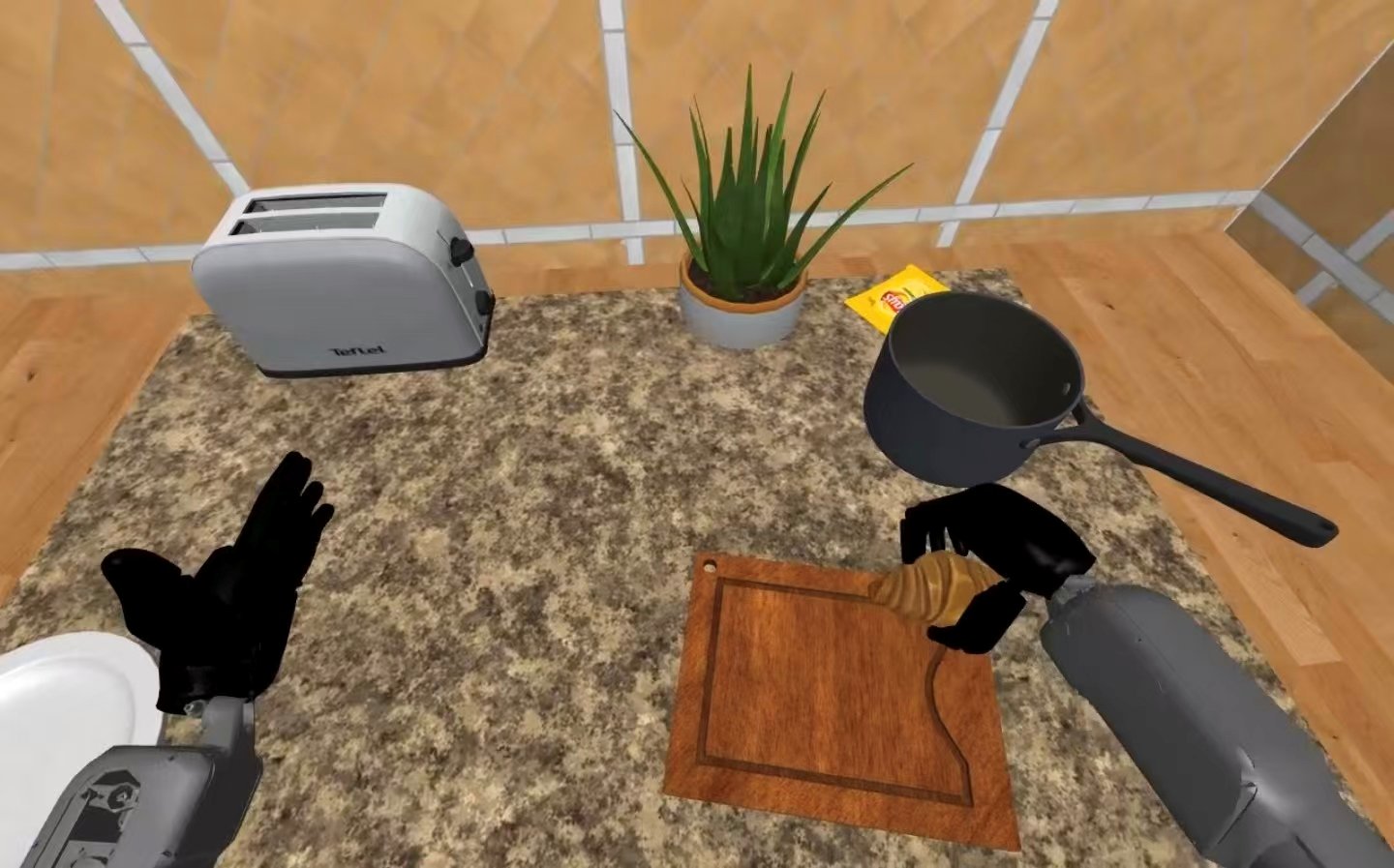}
    \end{minipage}
\hfill
\begin{minipage}{.15\textwidth}
        \centering
        \includegraphics[width=\textwidth]{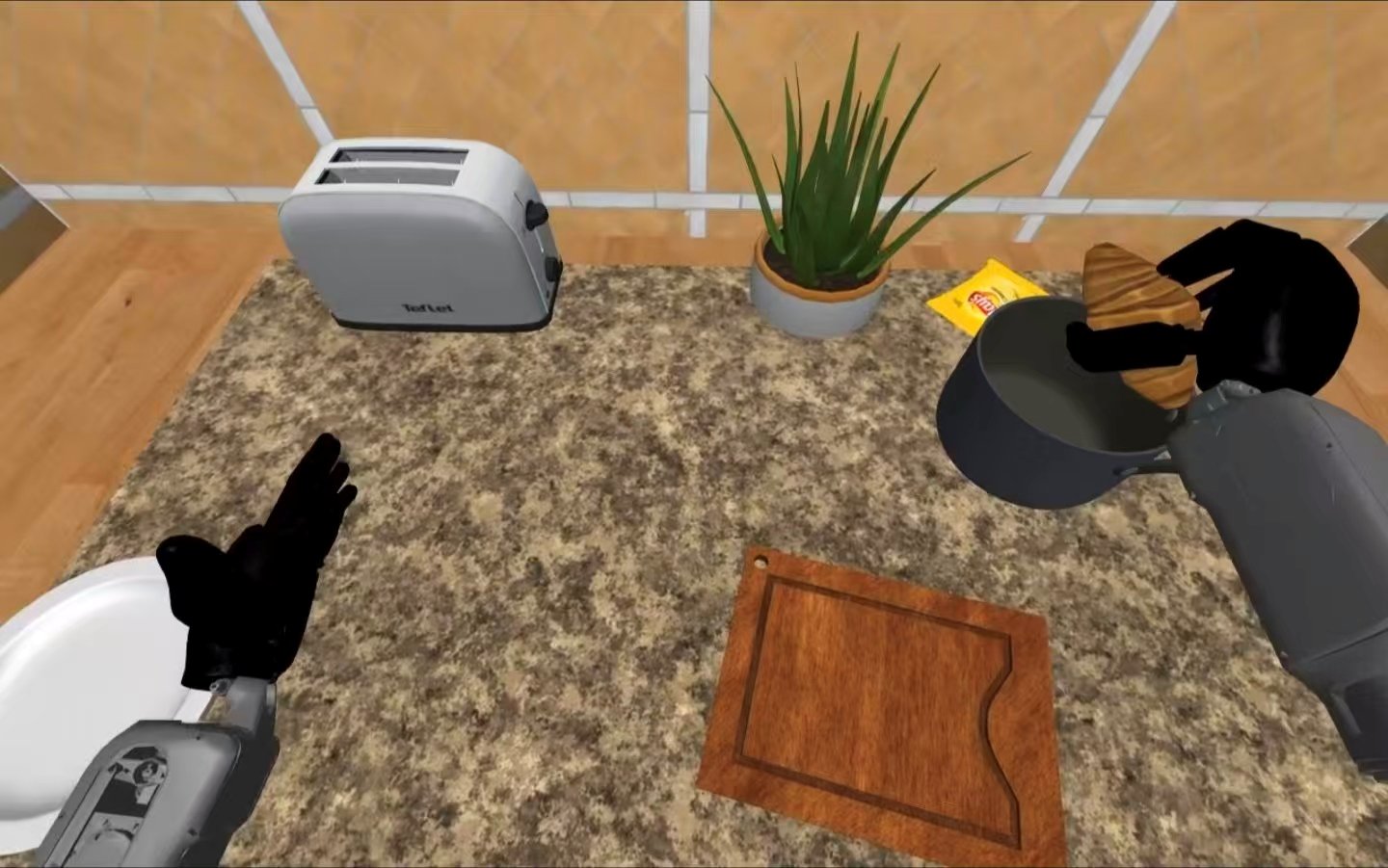}
    \end{minipage}
\hfill 
\begin{minipage}{.15\textwidth}
        \centering
        \includegraphics[width=\textwidth]{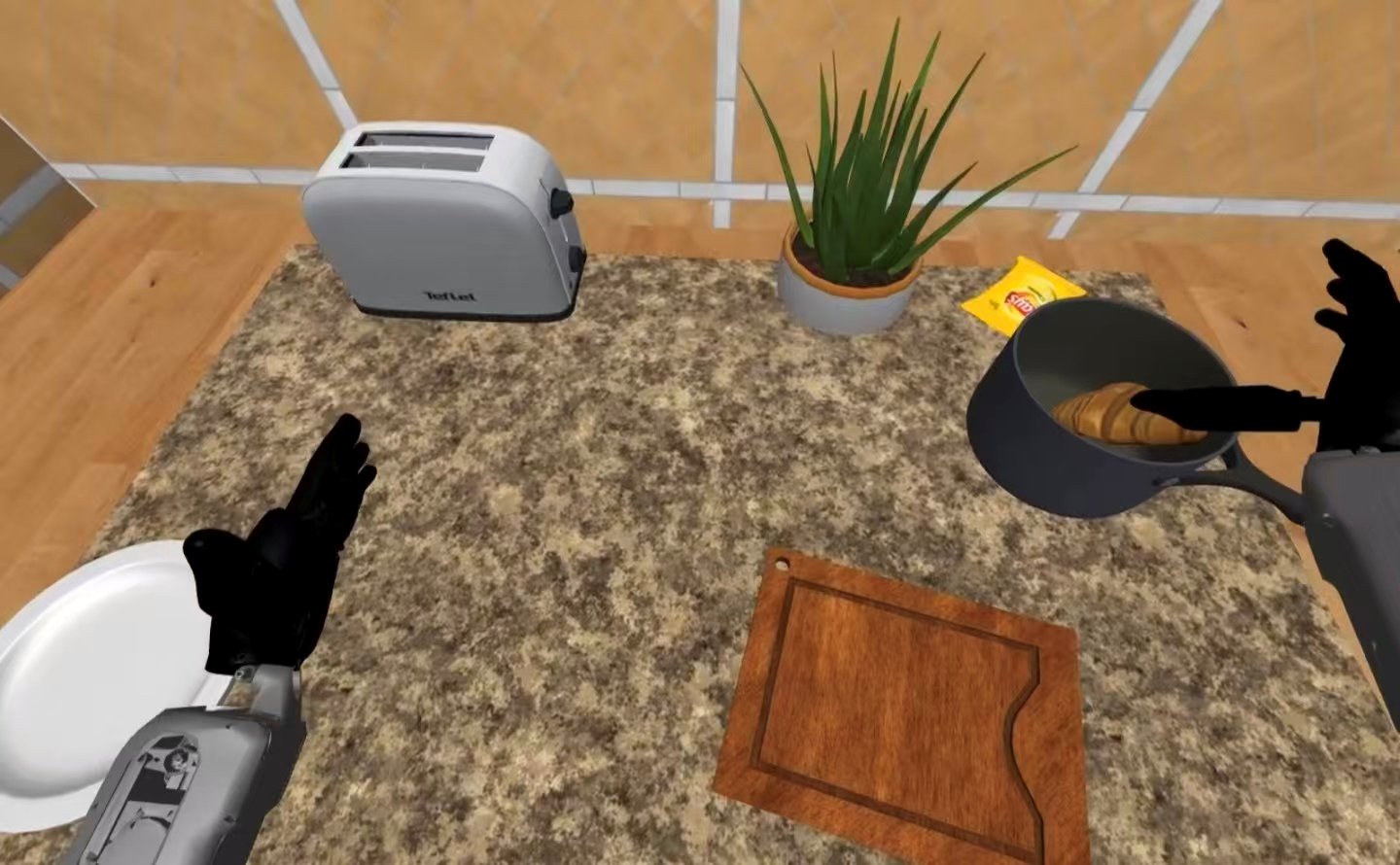}
    \end{minipage}
\hfill
\begin{minipage}{.15\textwidth}
        \centering
        \includegraphics[width=\textwidth]{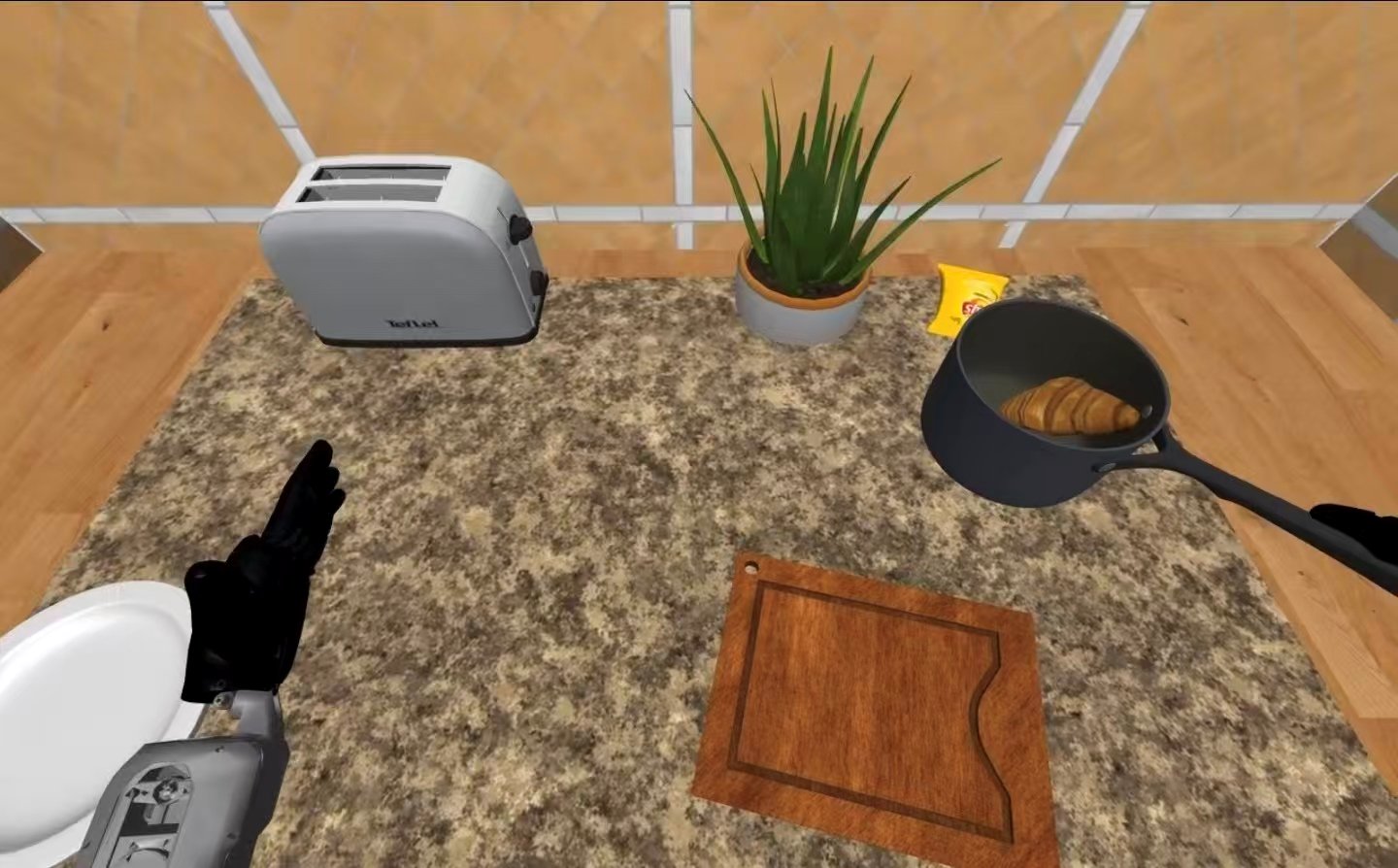}
    \end{minipage}
    
    % 第三组文字说明（左对齐）
    \raggedright
    \par\vspace{1em}
    AugVLA-3D: pick up the bread and put it in the pot
    \par\vspace{1em}
    \centering
    
    % 第四组图片
    
\begin{minipage}{.15\textwidth}
        \centering
        \includegraphics[width=\textwidth]{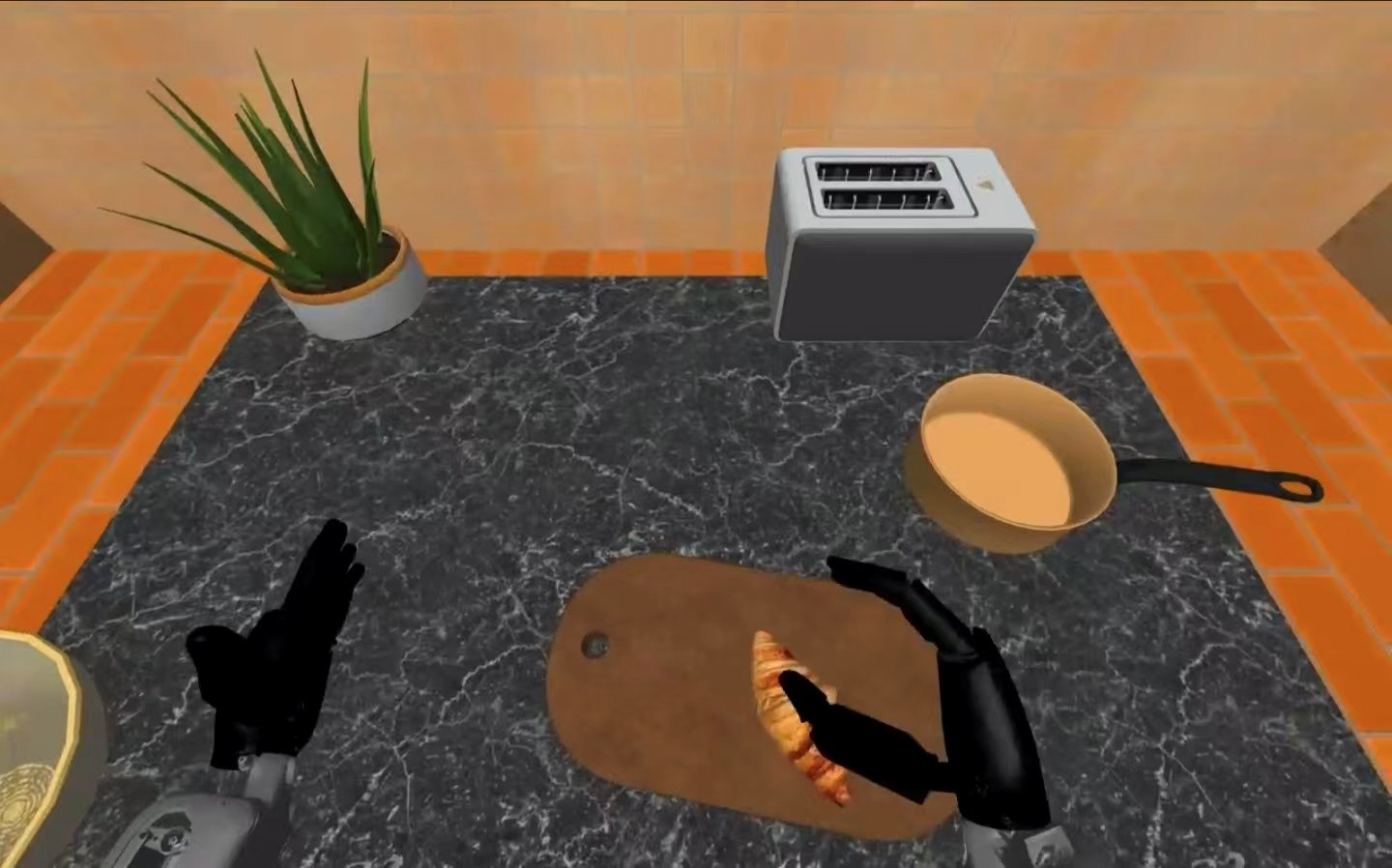}
    \end{minipage}
\hfill 
\begin{minipage}{.15\textwidth}
        \centering
        \includegraphics[width=\textwidth]{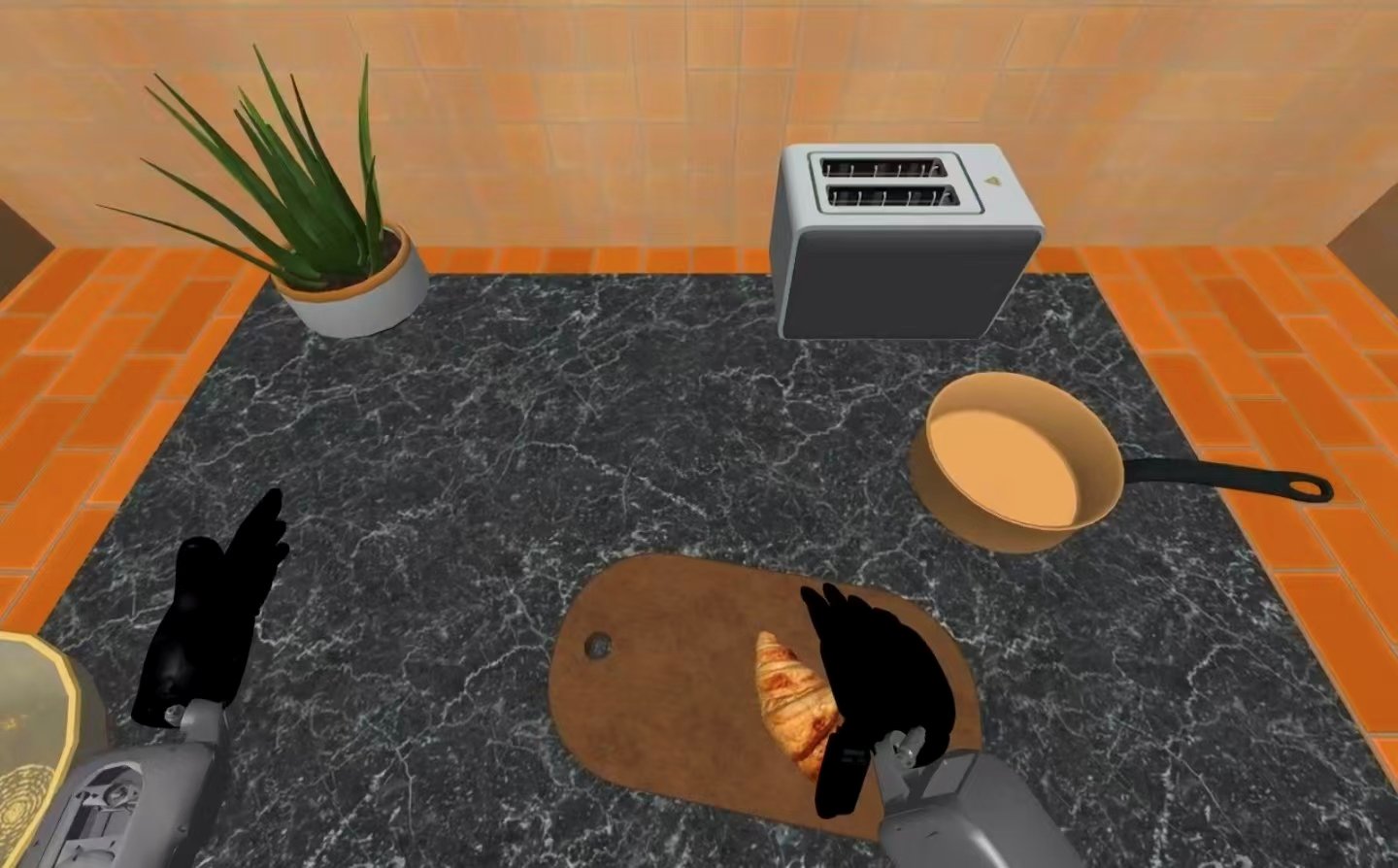}
    \end{minipage}
\hfill 
\begin{minipage}{.15\textwidth}
        \centering
        \includegraphics[width=\textwidth]{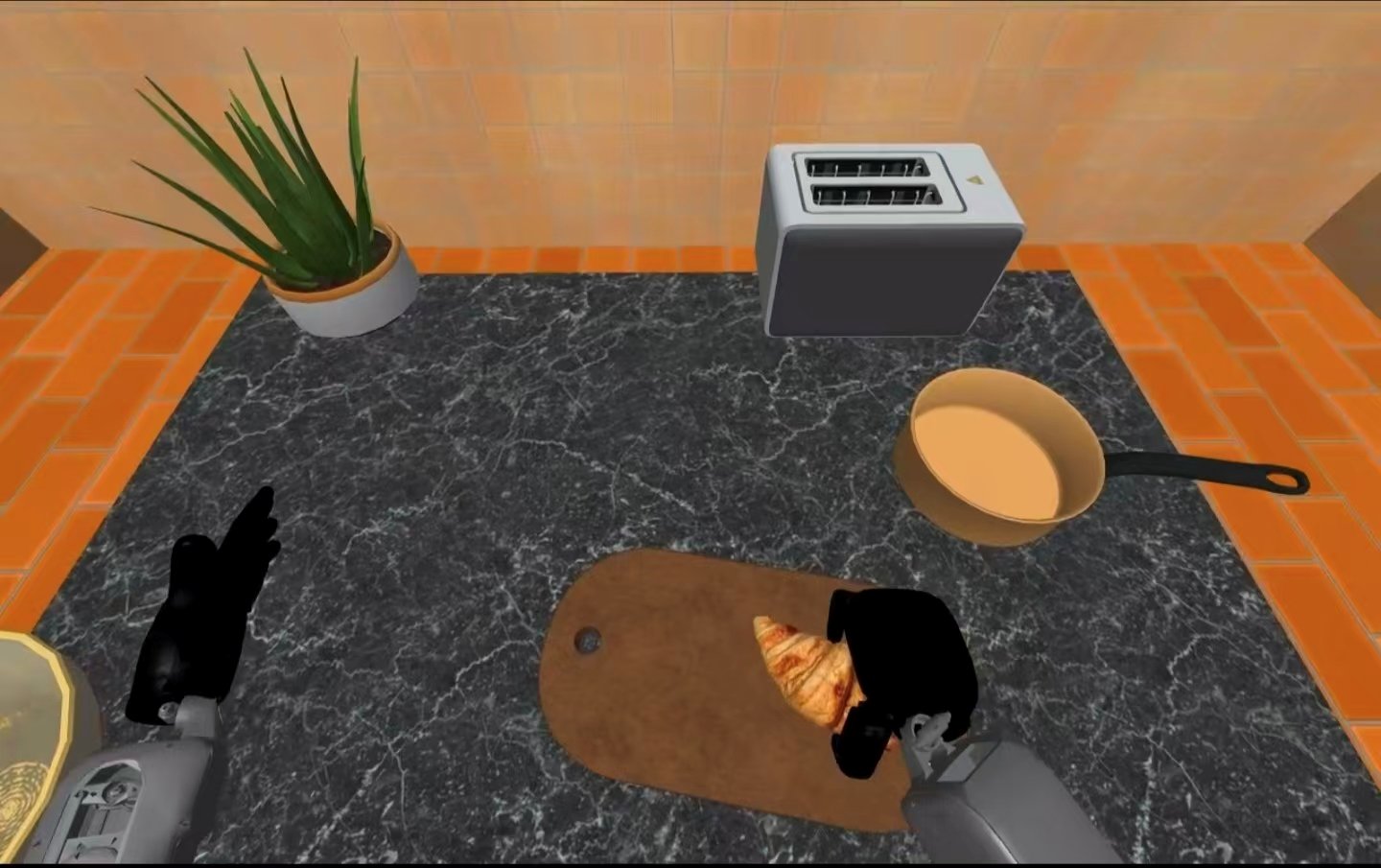}
    \end{minipage}
\hfill 
\begin{minipage}{.15\textwidth}
        \centering
        \includegraphics[width=\textwidth]{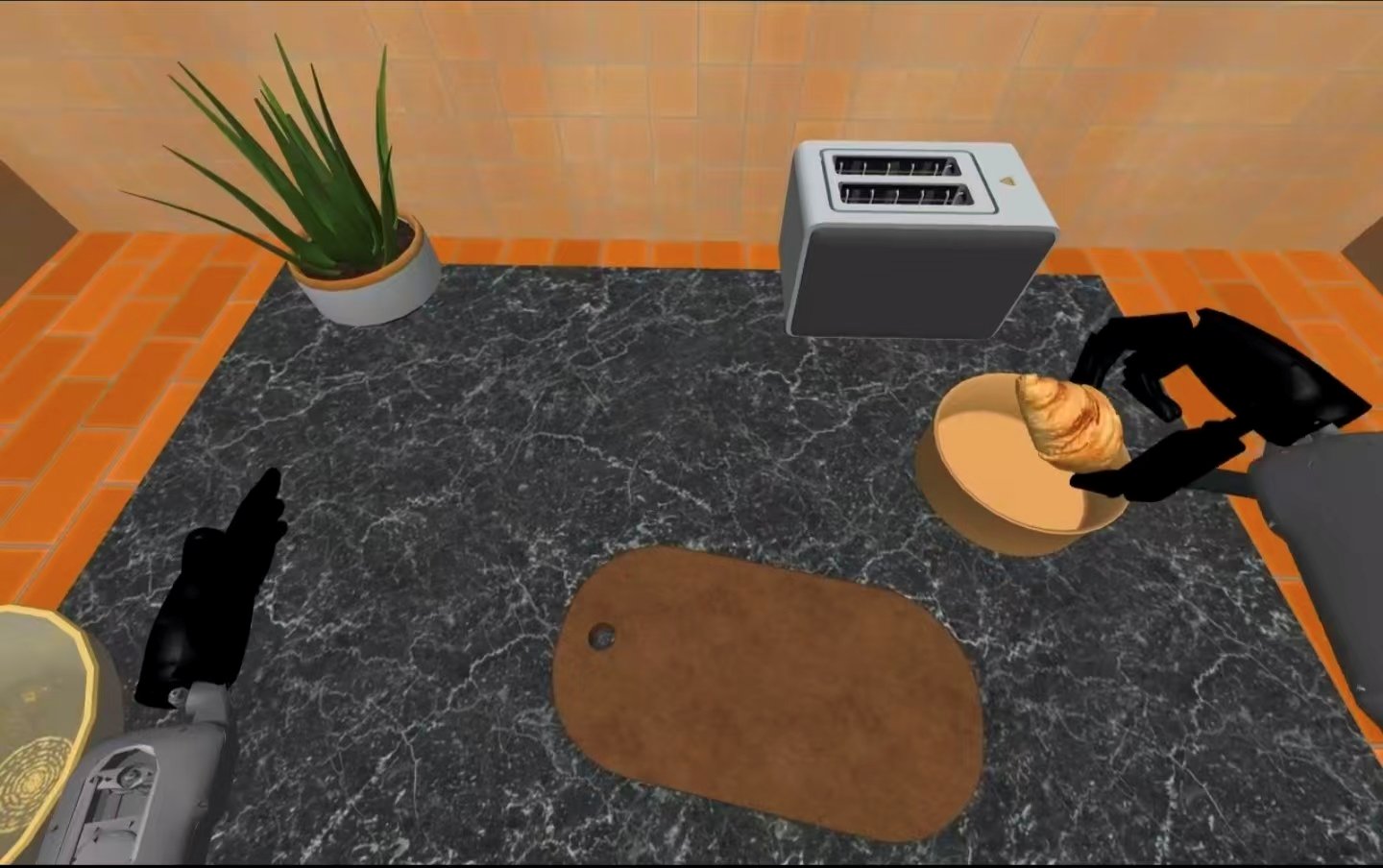}
    \end{minipage}
\hfill  
\begin{minipage}{.15\textwidth}
        \centering
        \includegraphics[width=\textwidth]{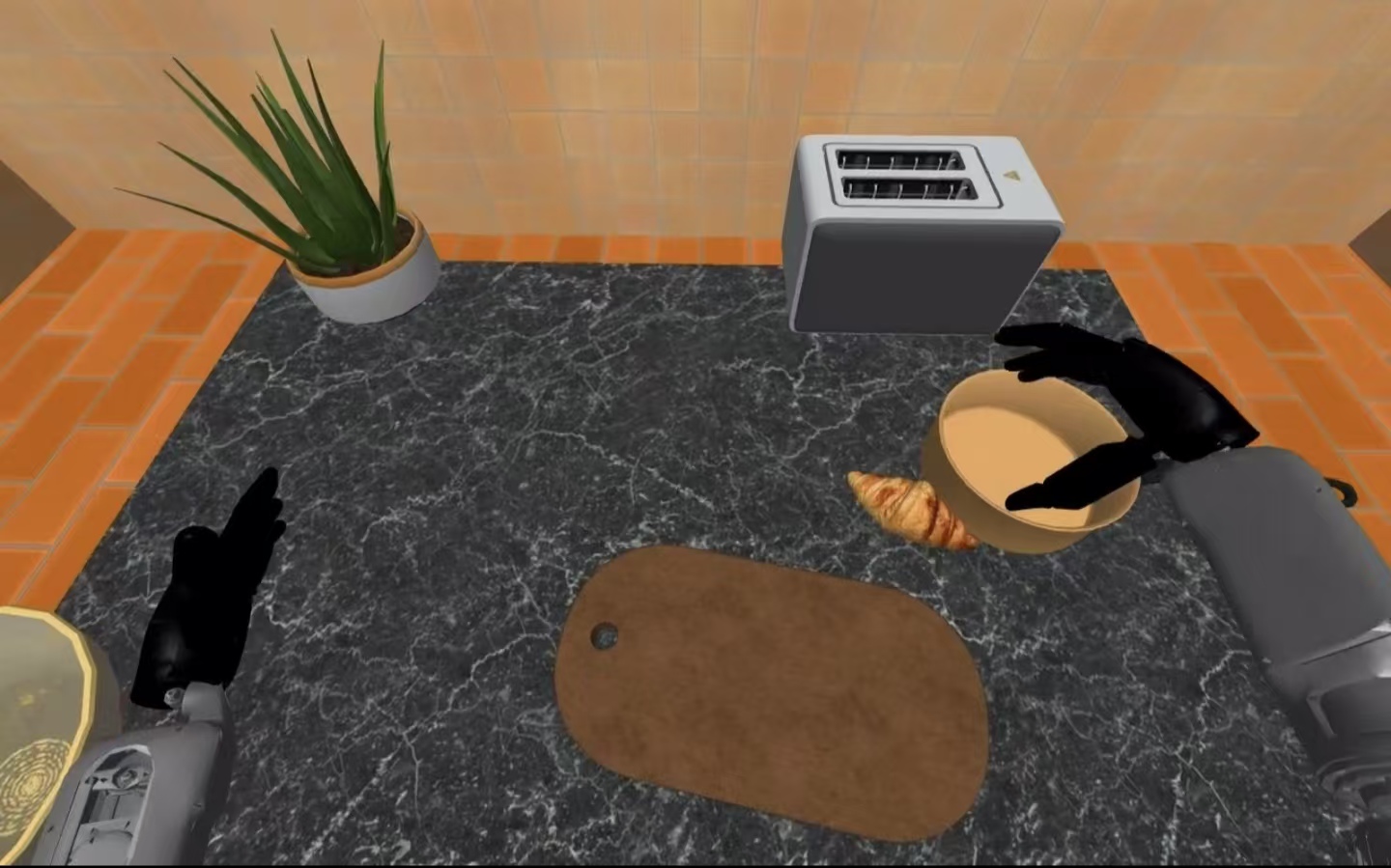}
    \end{minipage}
\hfill 
\begin{minipage}{.15\textwidth}
        \centering
        \includegraphics[width=\textwidth]{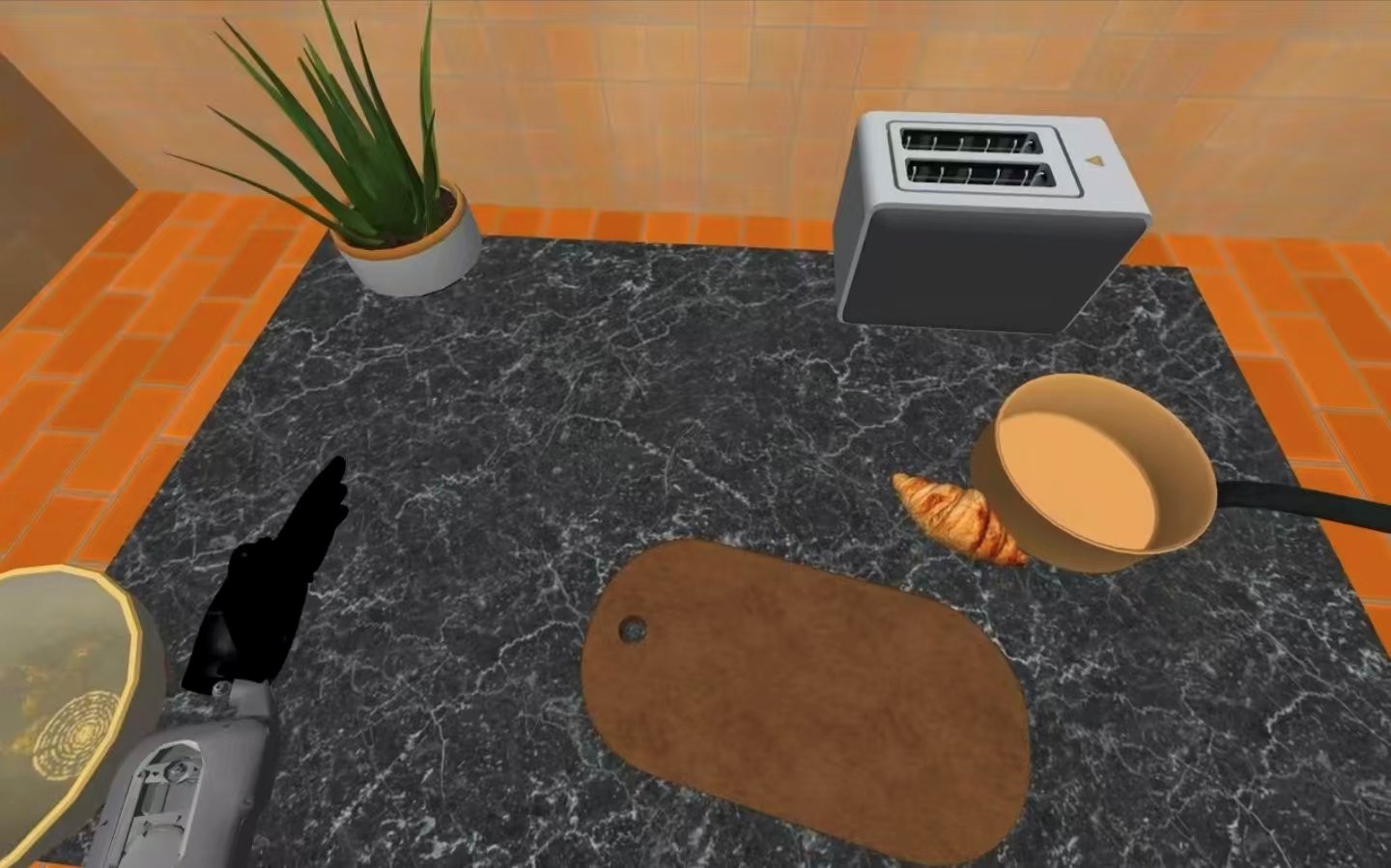}
    \end{minipage}
    
    % 第四组文字说明（左对齐）
    \raggedright
    \par\vspace{1em}
    Gr00T: pick up the bread and put it in the pot
    \par\vspace{1em}
    \centering
    
    \caption{Comparative experimental results between the AugVLA-3D and Gr00T models in complex manipulation scenarios with dexterous hands. The experiment involved two typical manipulation tasks: "pick up the cup and put it in a drawer, then close it" (rows 1-2) and "pick up a loaf of bread and put it in a pot" (rows 3-4). Each task consisted of six key action steps. To ensure fairness, the object layout, lighting conditions, and task instructions were kept consistent across all experimental scenarios. The results show that the AugVLA-3D model, which incorporates 3D spatial features, generally outperforms the Gr00T model in object positioning accuracy, motion trajectory smoothness, and task completion efficiency, validating the effectiveness of 3D features in improving robot manipulation intelligence.}
    \label{fig:all_images}
\end{figure*}
\textbf{Experimental Results.}
We evaluate our model against two state-of-the-art dexterous hand baselines, Diffusion Policy \cite{ze20243d} and Gr00T \cite{bjorck2025gr00t}. As shown in Fig.~\hyperref[figure:figure4]{4}, AugVLA-3D consistently outperforms both methods across multiple real-world scenarios. This performance gain stems from the explicit incorporation of 3D features, which equip the policy with a stronger awareness of geometric structure and spatial relations capabilities that are often underrepresented in conventional 2D-based VLAs. As a result, our model achieves more reliable grasping, placement, and coordination even under distractors or shifting object poses. Importantly, these advantages emerge despite training on highly constrained computational resources, suggesting that the integration of depth-driven reasoning provides an intrinsic robustness that cannot be easily compensated for by scale alone. We thus believe that with sufficient pretraining, the performance gap over existing baselines would widen further.

\subsection{Experimental Results on Simulation Benchmark}
We further benchmark our approach on \texttt{robocasa-gr1-tabletop-tasks}, which is built on the RoboCasa simulation framework and includes 24 official tabletop tasks designed for NVIDIA’s GR-1 humanoid foundation models. These tasks comprehensively assess the capacity of dexterous hands to operate across diverse manipulation contexts.

\textbf{Experimental Results.}
Following the GR00T setup, we train on 30 and 100 demonstrations per task, without any large-scale pretraining due to resource limitations. As summarized in Table~\hyperref[table:table1]{1}, AugVLA-3D achieves consistently higher success rates than both Diffusion Policy and Gr00T. This highlights the effectiveness of introducing 3D structural cues into VLA models, as they directly enhance spatial reasoning and object interaction fidelity: two bottlenecks for 2D-based approaches.

Qualitative results are shown in Fig.~\hyperref[figure:figure5]{5}. In the first task, which requires coordinated bimanual manipulation (“pick up the cup and place it into a drawer, then close it”), AugVLA-3D demonstrates superior performance, leveraging 3D feature injection to maintain consistent spatial alignment between the two hands and the manipulated objects. By contrast, Gr00T exhibits frequent miscoordination, reflecting its limited ability to model inter-hand dependencies in 3D space. In the second task (“pick up the bread and put it in the pot”), our model precisely estimates the relative geometry between the bread and the container, enabling successful placement. Gr00T, relying only on 2D features, often misjudges depth and distance, leading to failure. These findings underscore a broader insight: the introduction of explicit 3D features not only improves local grasp accuracy but also enhances global spatial reasoning, which is essential for complex, multi-object manipulations.

\section{Conclusion}

In this paper, we presented AugVLA-3D, a novel Vision-Language-Action framework that enhances conventional models with sensor-free 3D geometric features derived from 2D RGB inputs via monocular depth estimation, and further stabilized through an action-guided regularization module. This design substantially improves spatial reasoning, action prediction accuracy, and robustness in complex manipulation scenarios, outperforming state-of-the-art baselines such as Gr00T and Diffusion Policy in both real-world and simulation experiments. Importantly, our model achieves these gains despite being trained on limited resources, highlighting its inherent efficiency and scalability. Looking forward, large-scale pretraining, deployment to broader robotic platforms, and self-supervised depth refinement hold promise for further strengthening geometric fidelity and advancing the generalization of embodied AI systems.

\bibliographystyle{ieeetr}
% \bibliography{references}

\end{document}